\documentclass[conference]{IEEEtran}
\usepackage{times}
\usepackage[numbers]{natbib}
\usepackage{multicol}

\usepackage{amsmath,amsfonts,bm}

\def\eqref#1{equation~\ref{#1}}

\def\1{\bm{1}}

\DeclareMathAlphabet{\mathsfit}{\encodingdefault}{\sfdefault}{m}{sl}
\SetMathAlphabet{\mathsfit}{bold}{\encodingdefault}{\sfdefault}{bx}{n}

\DeclareMathOperator*{\argmax}{arg\,max}

\usepackage{graphicx}
\usepackage{xcolor}
\usepackage{tikz}
\usepackage{url}
\usepackage{float}
\usepackage{wrapfig}
\usepackage{etoolbox}
\makeatletter
\patchcmd{\@makecaption}{\normalfont\footnotesize\scshape}{\normalfont\footnotesize}{}{\PackageError{exrl}{Table caption patch failed}{}}
\makeatother
\usepackage{amsmath,amssymb,bbm}
\usepackage{booktabs,makecell}
\usepackage{algorithm}
\usepackage[endLComment=,italicComments=false]{algpseudocodex}
\usepackage[bookmarks=true]{hyperref}
\definecolor{exrlorange}{HTML}{C8753D}
\definecolor{exrlcurveorange}{HTML}{D95F02}
\hypersetup{colorlinks=true,allcolors=exrlorange,
 pdftitle={ExRL: Sample-Efficient Online Adaptation of Long-Chunk Policies via Execution Length Control},
 pdfauthor={Xu Luo, Jiaying Yang, Hao Wu, Junlin Xie, Zijin Hong, Ji Zhang, Lianli Gao, Jingkuan Song}}
\title{ExRL: Sample-Efficient Online Adaptation of\\Long-Chunk Policies via Execution Length Control}
\author{\authorblockN{Xu Luo$^{1}$, Jiaying Yang$^{2}$, Hao Wu$^{1}$, Junlin Xie$^{1}$, Zijin Hong$^{3}$, Ji Zhang$^{1}$, Lianli Gao$^{1}$, Jingkuan Song$^{2}$}
\authorblockA{$^{1}$UESTC\quad $^{2}$Tongji University\quad $^{3}$The Hong Kong Polytechnic University\\[3pt]
{\sublargesize\hypersetup{urlcolor=exrlcurveorange}\url{https://exrl-project.github.io}}}}
\renewcommand{\dbltopfraction}{0.74}
\IEEEoverridecommandlockouts
\IEEEaftertitletext{\vspace*{-11pt}
\begin{minipage}{\textwidth}
\centering
\includegraphics[width=\linewidth,trim=0 8.57bp 0 0,clip]{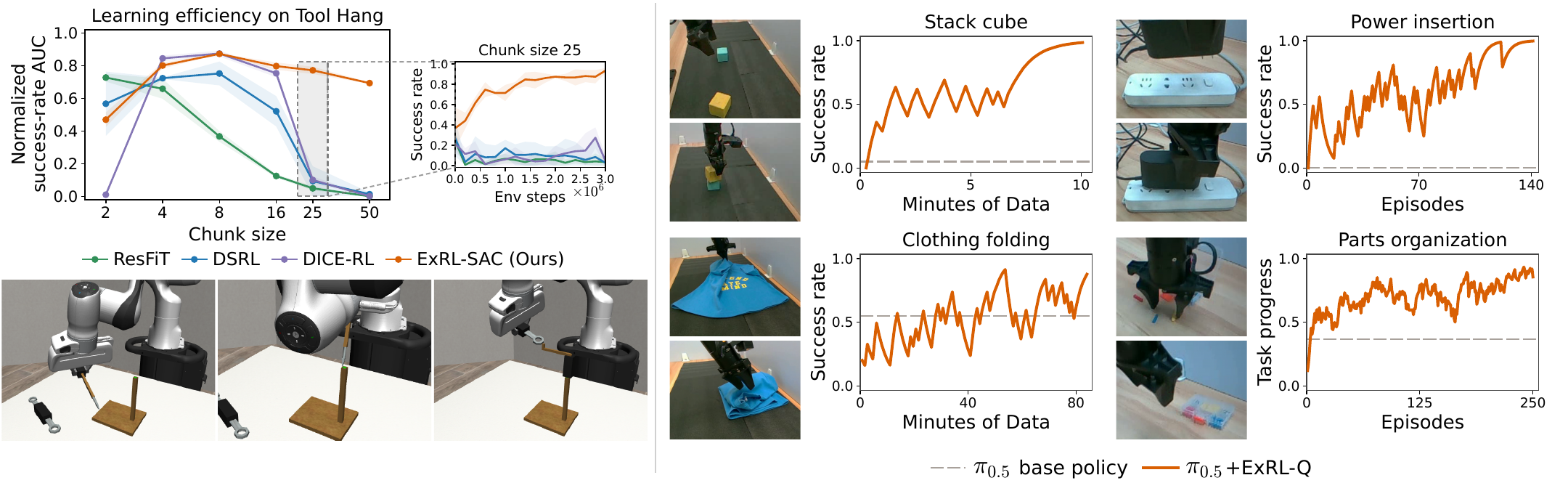}
\makeatletter\def\@captype{figure}\makeatother
\caption{\textbf{We propose ExRL, a sample-efficient RL method for adapting long-chunk policies through execution-length control.}
\emph{Left:} Existing RL methods lose sample efficiency as action chunks grow longer, while ExRL maintains high learning efficiency.
\emph{Right:} ExRL improves $\pi_{0.5}$ policies to high success rates on contact-rich and long-horizon real-world tasks with only tens of minutes of physical interaction or a few hundred episodes.}
\label{fig:intro}
\label{fig:real_world_adaptation}
\end{minipage}
\par\vspace{0.7\baselineskip}
}
\begin{document}

\maketitle
\makeatletter\global\@topnum\z@\makeatother

\begin{abstract}

Predicting long action chunks, often spanning dozens of timesteps, has become a key ingredient in increasingly capable robot policies. In this paper, we find that adapting these long-chunk policies with existing reinforcement learning (RL) methods can require many online samples due to the high dimensionality of the action chunks being optimized, making such adaptation impractical in real-world scenarios where interaction data are costly to collect. Our key insight is to shift adaptation from optimizing actions to optimizing how long to execute each predicted action chunk. This keeps the adaptation space low-dimensional while its expressivity grows with the chunk size. We thus introduce \emph{Execution at the Right Length} (ExRL), a simple approach which augments the base policy with a lightweight adapter that predicts the execution length of each chunk and is trained using off-policy RL algorithms. Across simulation benchmarks and real-world robotic tasks, ExRL enables sample-efficient adaptation of a diverse range of long-chunk policies with minimal task-specific hyperparameter tuning.

\end{abstract}

\section{Introduction}

Robot policies trained on demonstration datasets have enabled increasingly complex and generalizable behaviors across a wide range of robotic control tasks \citep{sunday2026act2preview,pi0.7}. A key technique shared by many of these policies is action chunking \citep{ACT}. Rather than predicting a single action at each query, action-chunked policies predict a sequence of actions and execute them open-loop, which can help capture non-Markovian structure in demonstrations \citep{lazzati2026does,liu2025bidirectional} and reduce compounding errors \citep{lazzati2026does,zhang2025action}. A common choice in recent robot policies is to predict \emph{long action chunks}, often exceeding 20 steps and approaching 50, as in vision-language-action (VLA) models such as $\pi_{0.5}$ \citep{pi0.5} and $\pi_0$-FAST \citep{fast}, and world action models such as DreamZero \citep{dreamzero}. By producing many actions per model query, long chunks enable high-frequency control (e.g., 50\,Hz) for dexterous tasks such as laundry folding \citep{pi_0}, while reducing inference pauses for smoother, faster execution.

These long-chunk robot policies, while performant, are rarely optimal when deployed in the real world due to limited demonstration coverage and imperfect generalization \citep{simchowitz2025pitfalls}. Reinforcement learning (RL) provides a direct way to adapt these policies  from online experience. In other domains such as language modeling, online RL has been shown to be quite effective for adapting models pretrained on offline datasets \citep{ouyang2022training,OpenAI-o1,DeepSeek-r1}. For robotic control, however, online data collection in the real world is often slow, costly, and sometimes unsafe, making sample efficiency a central requirement.

Yet we observe that long action chunks introduce a major bottleneck for sample-efficient adaptation (Figure~\ref{fig:intro}, left). RL methods that optimize entire action chunks or their latent variables face an increasingly high-dimensional optimization space as the chunk length grows. This substantially increases the amount of online experience required for adaptation.

This motivates us to seek an adaptation interface whose dimensionality does not grow with the chunk length. We observe that action-chunked policies naturally expose such an interface, namely, how long to execute a predicted chunk before querying the policy again. Although execution length is a simple scalar decision, adapting it to the current state and predicted action chunk can substantially alter policy behavior by controlling the trade-off between longer open-loop execution, which better exploits the benefits of action chunking, and shorter execution, which improves responsiveness to unexpected state changes or dynamics mismatch \citep{liu2025bidirectional}. This makes execution length a meaningful variable for policy improvement, while being particularly attractive for long chunks: as the chunk size grows, the number of possible execution lengths—and hence the expressivity—increases, whereas the optimization itself remains a low-dimensional discrete problem.

We instantiate this idea as \emph{Execution at the Right Length} (ExRL), a sample-efficient online RL method for adapting long-chunk policies.
It adds a lightweight execution-length policy that selects how many actions from each predicted chunk to execute before replanning, and trains it using standard off-policy RL algorithms. Because this adapter can be learned without modifying the base policy itself, ExRL provides a policy-agnostic interface that can be applied to virtually any class of action-chunked policies. Execution-length control is ultimately bounded by the capabilities of the underlying base policy. To go beyond this limitation, we can further optionally distill successful online trajectories back into the base policy during RL using simple imitation loss, allowing the base policy itself to improve alongside the adapter. Finally, ExRL can also leverage general offline data, such as human demonstrations, by reinterpreting arbitrary trajectory segments as off-policy experience under different execution lengths, further improving online sample efficiency.

We empirically demonstrate the effectiveness of ExRL across simulated and real-world manipulation tasks, showing state-of-the-art online adaptation performance across a range of long-chunk policies, largely without task-specific hyperparameter tuning. These include standard action-chunked policies, different classes of pretrained VLA models such as $\pi_{0.5}$ \citep{pi0.5} and $\pi_0$-FAST \citep{fast}, as well as a world action model, Cosmos Policy \citep{cosmos}. On real-world tasks spanning precise manipulation, long-horizon coordination, and clothing folding, ExRL efficiently adapts frozen $\pi_{0.5}$ policies, approaching perfect success on cube stacking with about ten minutes of physical interaction.

\section{Related Work}
\noindent\textbf{Imitation learning in robotics.} Imitation learning, particularly behavior cloning, has been widely used to train data-driven policies for robotic control \citep{atkeson1997robot,ross2011reduction,zhang2018deep,florence2022implicit,shafiullah2022behavior}. Prior work has explored a wide range of action parameterizations and policy architectures for imitation learning. Earlier methods often represent actions with Gaussian or GMM policies \citep{khansari2011learning,mandlekar2021matters}, while more recent approaches use expressive generative modeling techniques, including VAEs \citep{ACT}, diffusion or flow models \citep{DP,octo,pi_0,gr00t}, and autoregressive next-token prediction models \citep{rt-2,openvla,fast}. With the growing scale of human demonstration datasets and policy models \citep{bridge,droid,rt-1,oxe}, imitation learning has further become a pretraining paradigm for generalist robot policies, absorbing broad prior knowledge into a single model that can be adapted for real-world deployment \citep{octo,openvla,pi_0}. Our work provides an efficient method to adapt these diverse policy classes from online experience.

\noindent\textbf{Adaptive execution of action chunks.} Action chunking, as used in ACT \citep{ACT}, has become a de facto design choice in robot policies learned with imitation learning. Recognizing that the execution length of action chunks strongly affects policy behavior \citep{smolvla,zhang2025action,liu2025bidirectional}, several concurrent works have explored adaptive execution strategies. Earlier RL methods such as Dynamic Action Repetition \citep{lakshminarayanan2017dynamic} and TempoRL \citep{biedenkapp2021temporl} adapt how long a primitive action is repeated, rather than how much of a predicted action chunk is executed.
AAC \citep{AAC} selects the execution length of vision-language-action models using action entropy, while FFDC \citep{wang2026trust} selects the execution length for world action models using a learned verifier to check whether predicted future observations remain consistent with the real rollout. ExRL instead learns execution-length decisions directly from task rewards through online RL.
TAS \citep{weng2025temporal} uses online RL to select among actions cached from overlapping chunks at every timestep, rather than explicitly learning a chunk-level execution length.
Two other recent works, ACSAC \citep{ACSAC} and ACH \citep{shin2026adaptive},  learn value-based adaptive chunk-size selection for offline-to-online RL. However, these methods train and evaluate specific chunked policies within offline and offline-to-online RL frameworks, and are not designed as plug-in adapters for arbitrary pretrained action-chunked policies. The closest work to ours is \cite{zhao2026dynamic}, a concurrent study that also adapts action-chunked policies through reinforcement learning over execution length. However, it relies on PPO for optimization; our comparison in Appendix~\ref{app:ppo_comparison} shows that off-policy learning can adapt execution lengths with substantially fewer online samples.

\noindent\textbf{Reinforcement learning for robot policy adaptation.} On-policy methods such as PPO and GRPO have been used to fine-tune pretrained robot policies \citep{li2025simplevla,lu2025vla,chen2025pirl,dppo}, but often require many online rollouts, limiting their practicality on real robots. Off-policy RL algorithms such as SAC \citep{SAC} and TD3 \citep{TD3} can use data more efficiently, but their standard implementations typically use deterministic or Gaussian actors, making direct adaptation of more complex generative policies nontrivial. Policy-specific approaches address this challenge in different ways: DSRL \citep{DSRL} and ZPRL \citep{ZPRL} optimize diffusion/flow policies in latent spaces. OGPO \citep{ogpo} uses off-policy critics to guide optimization of the diffusion or flow generation process, while RL-100 \citep{rl-100} applies PPO-style updates across diffusion denoising steps. These methods can improve sample efficiency, but remain tied to particular policy classes. Residual RL instead learns additive corrections to the base actions \citep{Policy-decorator,ankile2025imitation,ankile2025residual,expo,DICE-RL}. While policy-agnostic and lightweight, residual RL still requires continuous action optimization; correcting an entire chunk makes this space grow with chunk length. RL Token \citep{rl-token} learns a lightweight actor conditioned on VLA representations and reference chunks, with regularization toward the base actions. PA-RL \citep{PA-RL} achieves policy-agnostic adaptation by casting RL as action optimization followed by supervised policy learning. RECAP \citep{pi0.6} builds an offline-to-online pipeline for training advantage-conditioned policies, which can incorporate batch online RL \citep{dong2025matters} for efficient adaptation.  In contrast to these action-modifying approaches, ExRL adapts the execution length of each predicted chunk, reducing online adaptation to a compact discrete decision space and enabling sample-efficient, policy-agnostic adaptation.

\suppressfloats[t]
\begin{figure}[t]
\centering
\includegraphics[width=\linewidth,trim=0 2.34bp 0 0,clip]{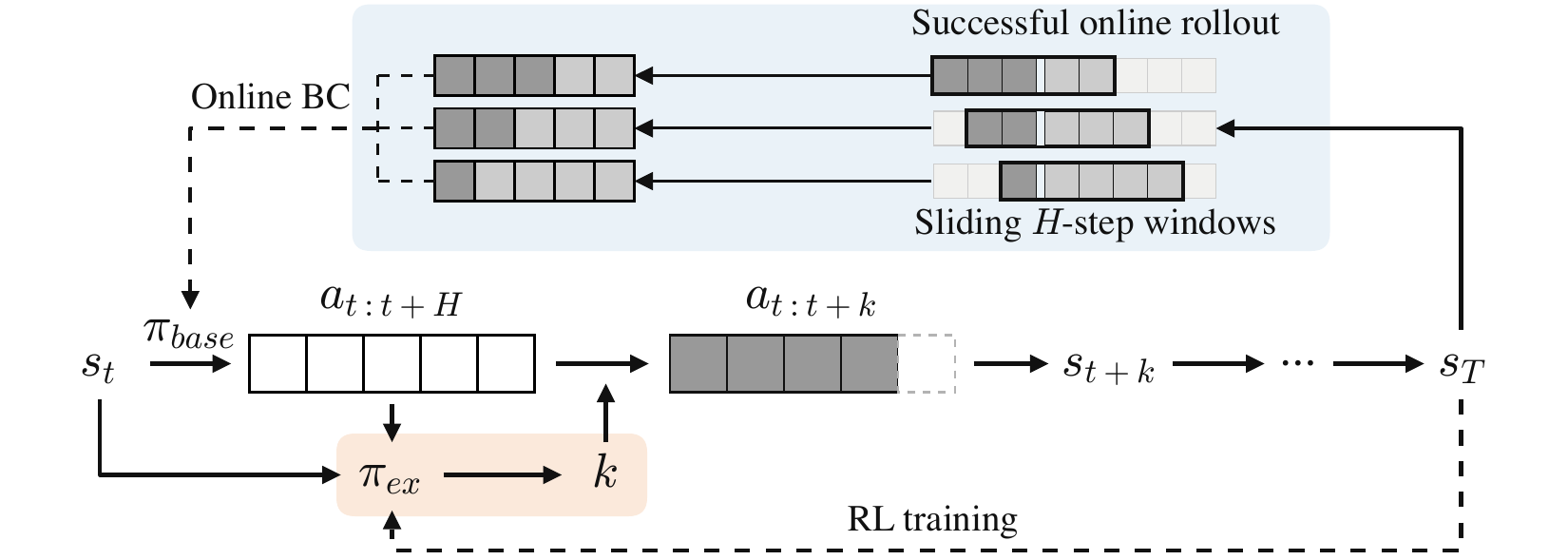}
\caption{ExRL learns execution lengths with RL, and we optionally use Online BC to distill successful online rollouts back into the base policy during RL.}
\label{fig:online_bc_method}
\end{figure}

\section{Preliminaries}
\label{sec:preliminaries}
\noindent\textbf{Markov decision processes.}
Markov decision process (MDP) is denoted by a tuple
$\mathcal{M}=(\mathcal{S},\mathcal{A},P,r,\rho_0,\gamma)$, where
$\mathcal{S}$ is the state space, $\mathcal{A}$ is the action space, $P(\cdot\mid s,a)$ is the transition dynamics,
$r:\mathcal{S}\times\mathcal{A}\rightarrow\mathbb{R}$ is the reward function,
$\rho_0$ is the initial state distribution, and $\gamma\in[0,1)$ is the discount
factor. A policy $\pi:\mathcal{S}\rightarrow\Delta(\mathcal{A})$ is a mapping from states to distributions over actions. The $Q$-function, or critic, of a policy $\pi$ is defined as
$Q^\pi(s,a)=\mathbb{E}_{\pi}[\sum_{t\ge0}\gamma^t r(s_t,a_t)\mid s_0=s,a_0=a]$,
which is the expected discounted return  of $\pi$ from taking action $a$ in
state $s$. The goal of reinforcement learning
is to find a policy that maximizes the expected discounted return.

\noindent\textbf{Q-learning.} A broad class of RL algorithms learns a parameterized Q-function $Q_\phi(s_t,a_t)$ by bootstrapping from Bellman targets. Given a transition $(s_t,a_t,r_t,s_{t+1})$, a generic one-step temporal-difference (TD) target takes the form \begin{equation} \label{eq:TD_target} y_t = r_t + \gamma V(s_{t+1}), \end{equation} where $V$ denotes the bootstrap value of the next decision state induced by the particular algorithm. The critic is then trained by minimizing the squared TD error \begin{equation}
\label{eq:TD_error} \min_{\phi}\; \mathbb{E}\left[ \left(Q_{\phi}(s_t,a_t)-y_t\right)^2 \right]. \end{equation}
\noindent\textbf{Action chunking.} An action-chunked policy
$\pi$ is a mapping $\mathcal{S}\rightarrow\Delta(\mathcal{A}^H)$ from states to distributions over $H$-step
action sequences.
At a query state $s_t$, a standard action-chunked policy samples an action chunk $a_{t:t+H}=[a_t,a_{t+1},...,a_{t+H-1}]\sim \pi(\ \cdot\mid s_t)$, and then
executes $a_{t:t+k}$, where $1\le k\le H$ is a fixed execution length, after which the policy is queried again at state $s_{t+k}$.

\noindent\textbf{Problem setting. } In this paper, we assume access to a pretrained action-chunked policy $\pi_{\mathrm{base}}$, and our goal is to adapt it to maximize the expected return specified by some reward $r$ using online interactions with an environment $\mathcal{M}$. Note that we make no assumptions on the parameterization or architecture of $\pi_{\mathrm{base}}$.

\section{Execution at the Right Length}
We propose ExRL, which adapts action-chunked policies by dynamically controlling how long each predicted action
chunk is executed before replanning, rather than modifying the action chunk
itself (Figure~\ref{fig:online_bc_method}). In this section, we formalize this idea as
reinforcement learning over execution-length decisions.

Given a pretrained action-chunked policy $\pi_{\mathrm{base}}$ with chunk size
$H$, we learn an additional execution-length policy $\pi_{\mathrm{ex}}:\mathcal{S}\times\mathcal{A}^H\rightarrow
\Delta(\mathcal{K})$, where $\mathcal{K}=\{1,\ldots,H\}$.
At each query, $\pi_{\mathrm{base}}$ first predicts an action chunk, and $\pi_{\mathrm{ex}}$ maps the
current state and this predicted chunk to a distribution over feasible execution
lengths. In this way, the raw
primitive actions are still produced by $\pi_{\mathrm{base}}$, while the learned execution-length
policy acts in the much smaller discrete space $\mathcal{K}$, avoiding direct optimization in the continuous,
high-dimensional space of action chunks. We can also optionally allow the policy
to choose $k=0$, corresponding to executing none of the actions in the current
chunk, which is essentially an immediate replanning decision.

\noindent\textbf{Duration-aware Q-learning.}
The execution-length decision naturally leads to a temporally extended
Q-function. At a query timestep $t$, let $a_{t:t+H}$ denote the action chunk sampled from
$\pi_{\mathrm{base}}(\ \cdot\mid s_t)$. In ExRL, we learn a critic
$Q_\phi(s_t,a_{t:t+H},k)$, which estimates the return obtained by executing the first
$k$ actions of the predicted chunk and then continuing with the current
execution-length strategy. This is closely related to Q-chunking \citep{li2026reinforcement}, which
learns a critic over an action sequence rather than a single primitive action. The difference is that in Q-chunking, the critic evaluates the action chunks, while the critic in ExRL evaluates the selected prefix lengths of action chunks. Thus, the accumulated reward and the
bootstrap discount in ExRL both depend on the chosen ``meta-action'' $k_t$, leading to the duration-aware TD target below:
\begin{equation}
    \label{eq:duration_td_target}
    y_t
    =
    \sum_{i=0}^{k_t-1}\gamma^i r(s_{t+i},a_{t+i})
    +
    \gamma^{k_t}
    V(s',a'),
\end{equation}
where $s' = s_{t+k_t}$ is the next query state, $a'=a_{t+k_t:t+k_t+H}\sim\pi_{\mathrm{base}}(\ \cdot\,\mid s')$ is the next chunk sampled from the frozen base policy, and $V$ is the next-state value. This duration-aware backup mirrors the Bellman backup used in semi-Markov
decision processes (SMDPs)~\citep{SMDP}, where temporally extended decisions may
last for a variable number of primitive timesteps. In Appendix~\ref{app:smdp}, we
formalize this connection by showing how execution-length control induces an
SMDP under the frozen base policy $\pi_{\mathrm{base}}$.

\noindent\textbf{Practical instantiations.}
ExRL is agnostic to the choice of RL algorithm, as long as the algorithm can
optimize over the discrete execution-length space with the duration-aware backup
above. In practice, we instantiate ExRL with two off-policy variants for
sample-efficient online adaptation. The first variant, \textbf{ExRL-Q}, utilizes the discrete nature of the action space and
uses value-based RL without learning an explicit actor \citep{watkins1992q,mnih2013playing}. In this case, the next-state value
$V$ is computed by a Bellman optimality backup:
\begin{equation}
    V(s',a')
    =
    \max_{k'\in\mathcal{K}} Q_{\bar{\phi}}(s',a',k'),
\end{equation}
where $Q_{\bar{\phi}}$ is a  slowly-updated target, and the critic is trained by minimizing the squared TD error:
\begin{equation}
    \mathcal{L}_Q(\phi)
    =
    \mathbb{E}
    \left[
        \left(
            Q_\phi(s_t,a_{t:t+H},k_t)-y_t
        \right)^2
    \right].
\end{equation}
During online rollout, the execution
length is selected greedily as
\begin{equation}
    k_t
    =
    \argmax_{k\in\mathcal{K}} Q_\phi(s_t,a_{t:t+H},k).
\end{equation}

\suppressfloats[t]
\begin{figure}[t]
\centering
\includegraphics[width=\linewidth,trim=0 2.34bp 0 0,clip]{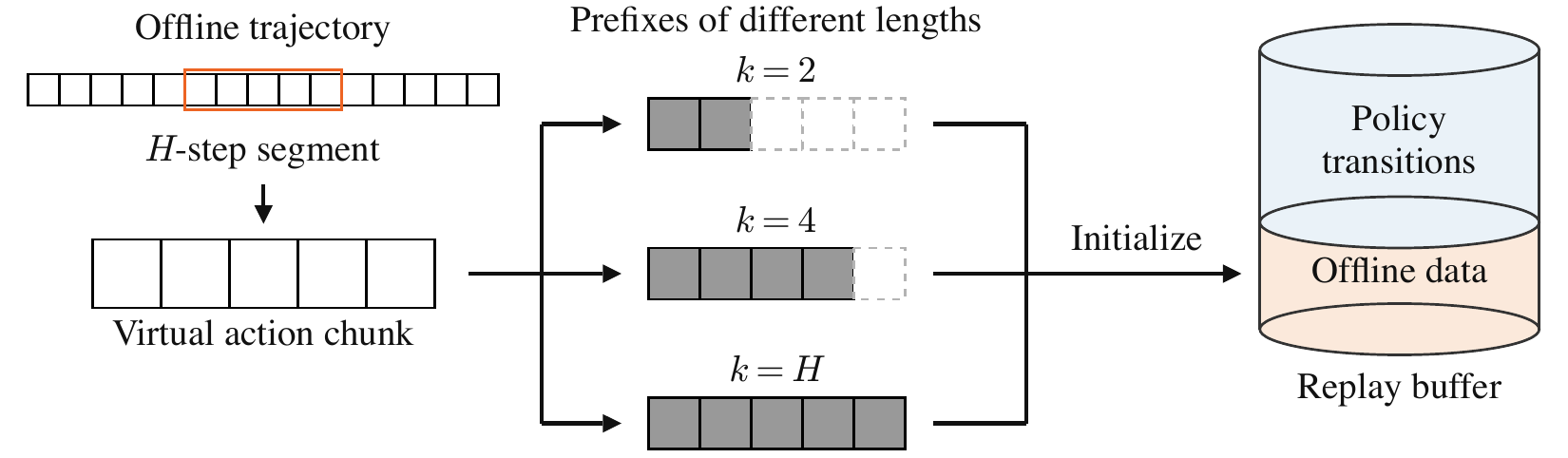}
\caption{ExRL can also leverage offline data by treating each $H$-step trajectory segment as a virtual action chunk, whose prefixes provide replay experience for different execution lengths.}
\label{fig:offline_data_method}
\end{figure}

The second variant, \textbf{ExRL-SAC}, adds an explicit actor
$\pi_\psi(k_t\mid s_t,a_{t:t+H})$ that outputs a categorical distribution over
execution lengths. Since $\mathcal{K}$ is discrete, we use discrete SAC
\citep{christodoulou2019soft} as an actor-critic instantiation. Compared
with continuous-action SAC, the discrete action space allows us to compute the
policy value exactly by summing over all $k\in\mathcal{K}$. In this case, the
hard maximum in ExRL-Q is replaced by the soft value under the actor. The soft value is
\begin{equation}
\begin{aligned}
    V(s',a') ={}& \sum_{k'\in\mathcal{K}} \pi_\psi(k'\mid s',a')\bigl[
        Q_{\bar{\phi}}(s',a',k')\\
        &\hspace{2em}-\alpha\log\pi_\psi(k'\mid s',a')\bigr].
\end{aligned}
\end{equation}
The entropy term $-\alpha\log\pi_\psi$ encourages the policy to maintain
uncertainty over execution lengths, which improves exploration and prevents
early collapse to a single fixed $k$. Squared TD error is used to update the critic, and the actor is trained to choose execution
length with high soft value:
\begin{equation}
\begin{aligned}
    \mathcal{L}_\pi(\psi) ={}& -\mathbb{E}_{k_t\sim\pi_\psi(\cdot\mid s_t,a_{t:t+H})}\bigl[
        Q_\phi(s_t,a_{t:t+H},k_t)\\
        &\hspace{2em}-\alpha\log\pi_\psi(k_t\mid s_t,a_{t:t+H})\bigr].
\end{aligned}
\end{equation}

Both ExRL-Q and ExRL-SAC further use conservative Q-learning
(CQL) \citep{CQL} to reduce overestimation of
execution lengths with limited support in the replay data.
Specifically, we augment the TD objective with
\begin{multline}
\label{eq:critic_cql}
\mathcal{L}_{\mathrm{critic}}(\phi)=\mathcal{L}_Q(\phi)
+\lambda_{\mathrm{CQL}}\mathbb{E}_{(s,a,k)\sim\mathcal{D}}\Biggl[\\
\log\biggl(\frac{1}{|\mathcal{K}|}
\sum_{k'\in\mathcal{K}} \exp Q_\phi(s,a,k')\biggr)
-Q_\phi(s,a,k)\Biggr],
\end{multline}
where $\mathcal{D}$ is the replay buffer and $\lambda_{\mathrm{CQL}}$ controls the
regularization strength. The algorithms of ExRL-Q and ExRL-SAC are summarized in Algorithm \ref{alg:exrl_q_full} and \ref{alg:exrl_sac_full}, respectively.

\begin{figure*}[t]
      \centering
      \includegraphics[width = 1.0\linewidth,trim=0 10.97bp 0 0,clip]{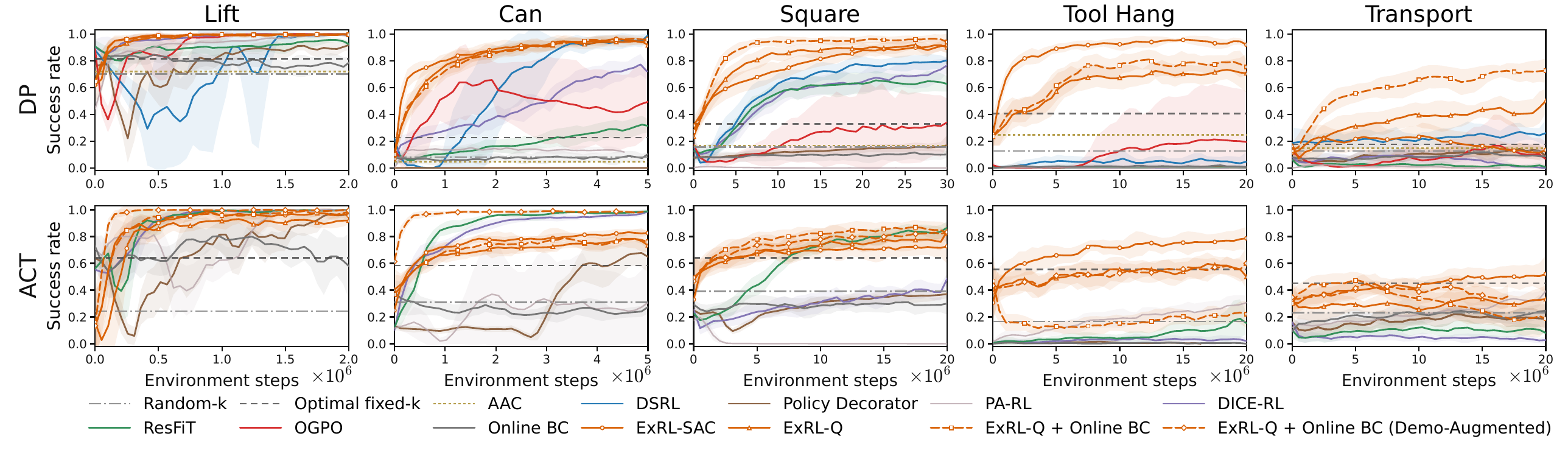}
      \caption{ExRL enables sample-efficient online adaptation of DP and ACT of chunk size $50$ on \texttt{Robomimic}.}
      \label{fig:robomimic_comparison}
\vspace{4pt}
      \centering
      \includegraphics[width = 1.0\linewidth,trim=0 12.4bp 0 0,clip]{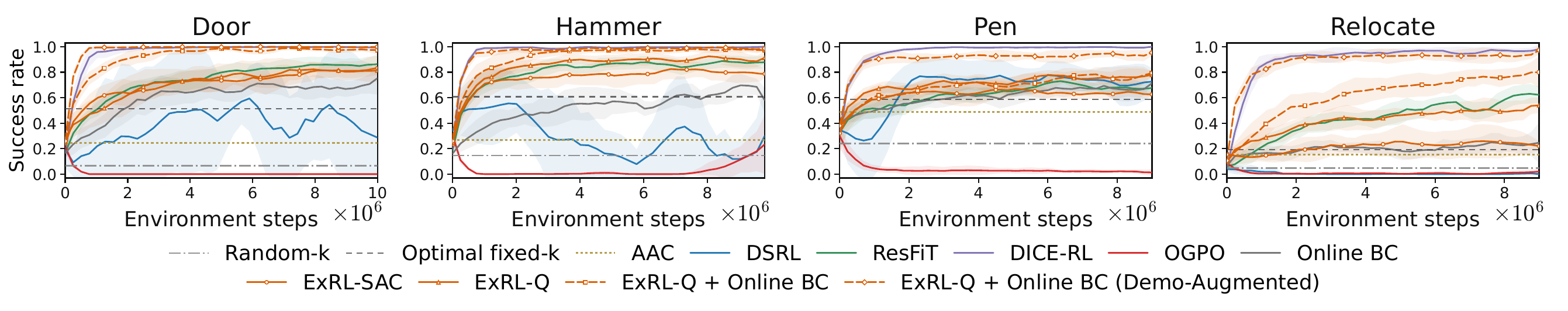}
      \caption{Online adaptation of DP on \texttt{Adroit} with chunk size $H=32$.}
      \label{fig:adroit_comparison}
\end{figure*}

\noindent\textbf{ExRL + online behavior cloning (BC).}
ExRL can improve policies without changing the actions proposed within each chunk, but its capability is ultimately bounded by the underlying base policy. How can we go beyond this limitation? We note that, by replanning at adaptive times, ExRL can stitch together prefixes from different predicted chunks and produce successful trajectories that would not arise from simply executing the original long chunks open-loop. We can therefore distill these improved behaviors back into the base policy during RL (Figure~\ref{fig:online_bc_method}). Specifically, from each successful ExRL rollout, we construct $H$-step action-chunk targets from the primitive actions actually executed, even when a target spans multiple replanning decisions, and update $\pi_{\mathrm{base}}$ using the same behavior cloning objective as in pretraining. In this way, behaviors discovered through execution-length adaptation become direct supervision for the base policy, improving its chunk predictions and in turn providing a stronger basis for further adaptation.

\noindent\textbf{Leveraging offline data.}
ExRL can also reuse general offline trajectories, such as human demonstrations, even if they were collected without action chunking. The key observation is that any sequence of recorded primitive actions can be retrospectively treated as a virtual action chunk (Figure~\ref{fig:offline_data_method}). Its prefixes of different lengths then provide real experience of executing the same action sequence for different lengths. Concretely, before online training, we enumerate trajectory segments of all lengths $k \leq H$ from offline trajectories and add the corresponding execution-length transitions to the replay buffer. For each transition, the offline trajectory provides the accumulated rewards and resulting state for the executed prefix; from that state onward, the value target bootstraps with the current base policy and execution-length strategy. 
This anchors the value of each virtual chunk to the expected return under ExRL's actual execution strategy, rather than to the behavior that generated the offline trajectory. When online behavior cloning is enabled, demonstration data can also be mixed with successful online trajectories to jointly update the base policy.

\noindent\textbf{Remark.} One may ask why we do not simply use a short fixed execution length for previous RL methods to avoid high-dimensional optimization, or first select the optimal fixed execution length and then optimize on top of it. Always using a short execution length for RL, however, can be problematic for several reasons: it may yield very poor initial performance on some tasks, making RL optimization difficult; it requires more frequent policy queries, increasing training cost and inference overhead while potentially introducing inference--execution synchronization issues; and, on long-horizon tasks, it substantially increases the effective decision horizon, further complicating RL. Alternatively, preselecting an optimal fixed execution length instead requires extensive real-world evaluation, potentially hundreds of rollouts, which is already comparable to the interaction budget of a typical real-world online RL experiment.

\section{Experiments}
The goal of our experiments is to answer the following core questions:
\begin{enumerate}
\renewcommand{\labelenumi}{(\textbf{Q\arabic{enumi}})}
    \item Can ExRL enable sample-efficient adaptation across action-chunked policies, including generalist robot policies in both simulation and real-world tasks?
    \item Can updating the base policy with successful online trajectories and demonstrations further improve adaptation?
    \item What execution-length patterns does ExRL learn?
\end{enumerate}

\subsection{Base Policies}
To show the generality of our method, we conduct experiments on diverse classes of action-chunked policies. We consider both representative action-chunked policies trained with imitation learning and several pretrained generalist robot policies of different classes.

\noindent\textbf{Representative action-chunked policies}:
(1) \textbf{Diffusion Policy (DP)}~\citep{DP}: an action-chunked policy that models multimodal continuous action chunks using denoising diffusion~\citep{ho2020denoising};
and (2) \textbf{ACT}~\citep{ACT}: a transformer-based policy that predicts action chunks using an encoder--decoder architecture.

\noindent\textbf{Generalist robot policies.}
(1) $\pi_0$~\citep{pi_0} and $\pi_{0.5}$~\citep{pi0.5}: state-of-the-art VLA models that use flow matching \citep{lipman2022flow} to model action chunks;
(2) \textbf{$\pi_0$-FAST}~\citep{fast}: an autoregressive VLA model that generates action chunks through efficient action tokenization;
and (3) \textbf{Cosmos Policy}~\citep{cosmos}: a world action model built on a large pretrained video generation model.

\begin{figure*}[t]
      \centering
      \includegraphics[width = 1.0\linewidth,trim=0 12.4bp 0 0,clip]{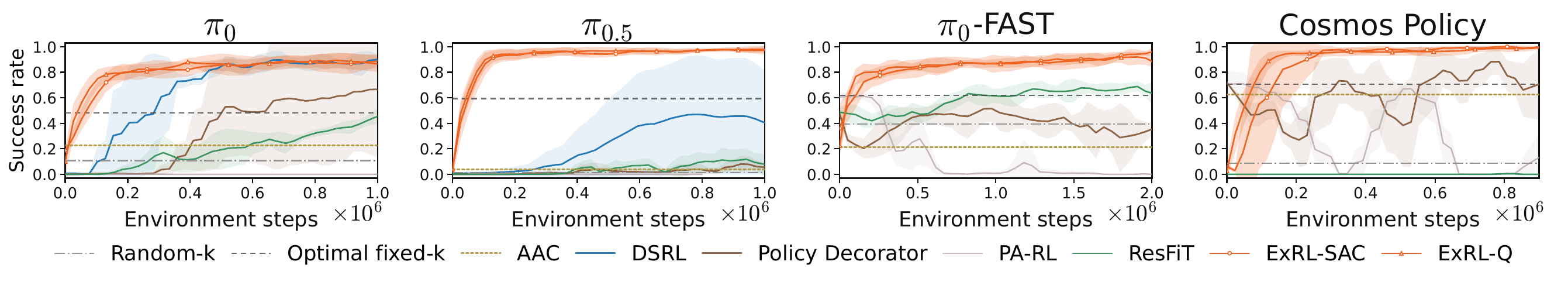}
      \caption{ExRL enables sample-efficient online adaptation of different classes of generalist robot policies on \texttt{Libero}, including the flow-based VLA models $\pi_0$ and $\pi_{0.5}$, the autoregressive VLA model $\pi_0$-FAST, and the world action model Cosmos Policy. All base policies are of chunk size $50$.}
      \label{fig:libero_comparison}
\vspace{4pt}
\edef\savedfigurecounter{\arabic{figure}}
\setcounter{figure}{7}
      \centering
      \includegraphics[width = 1.0\linewidth,trim=0 1.85bp 0 0,clip]{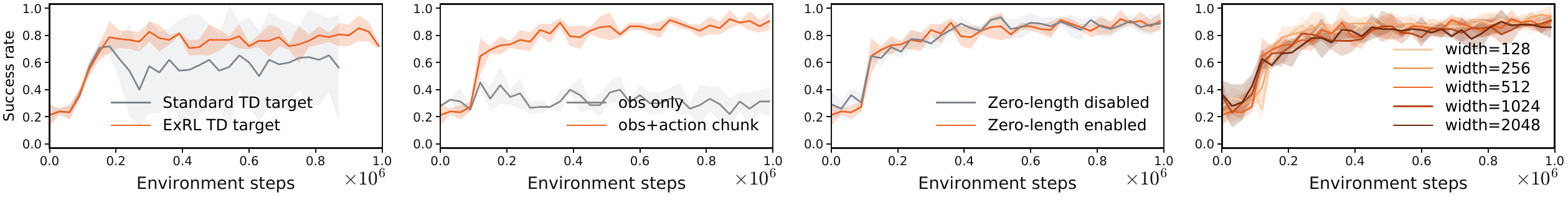}
     \caption{\textbf{Ablation studies on \texttt{Can} with $H=32$.} Left to right: TD target, conditioning on the predicted action chunk, zero-length execution, and network width.}
      \label{fig:other_ablations}
\setcounter{figure}{\savedfigurecounter}
\end{figure*}

\begin{figure*}[t]
\edef\savedanalysisfigurecounter{\arabic{figure}}
\setcounter{figure}{8}
      \centering
      \includegraphics[width=\linewidth,trim=0 9.52bp 0 0,clip]{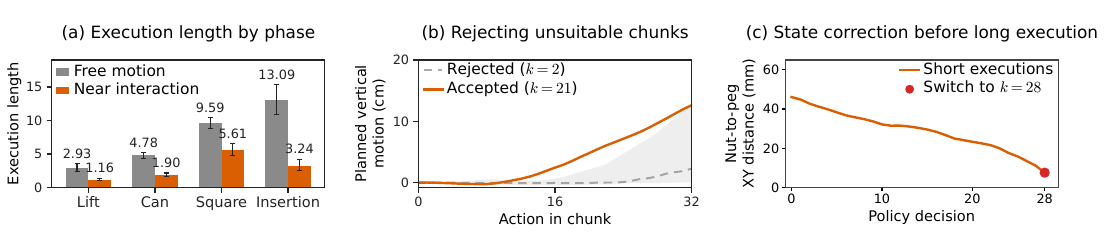}
      \caption{\textbf{How ExRL adapts execution length.}
(a) Mean execution lengths during free motion and near interaction, with 95\% trajectory-bootstrap intervals. (b) Planned vertical motion over rejected and accepted full chunks on \texttt{Square}; gray shows the median and 10--90\% range of rejected proposals. (c) In another \texttt{Square} rollout, short executions progressively align the nut with the peg before ExRL selects a long execution.}
      \label{fig:visulization}
\setcounter{figure}{\savedanalysisfigurecounter}
\end{figure*}

\subsection{Baselines and Setup}
We compare ExRL against execution-length baselines and several online adaptation baselines, with a focus on policy-agnostic methods for action-chunked policies. Our baselines include:

\noindent\textbf{Execution-length baselines.}
(1) \textbf{Random-$k$}: randomly samples an execution length from the feasible set at each policy query;
(2) \textbf{Optimal fixed-$k$}: sweeps fixed execution lengths and reports the best-performing one.\\
\noindent\textbf{AAC}~\citep{AAC}: A training-free adaptive execution method that selects the execution length using action entropy.\\
\noindent\textbf{Online BC}: An ablation that directly distills successful online rollouts back into the base policy using behavior cloning without RL.\\
\noindent\textbf{Policy Decorator}~\citep{Policy-decorator}: An off-policy residual RL method that freezes the base policy and learns additive corrections to its predicted actions, with bounded residuals and controlled exploration for stable online learning.\\
\noindent\textbf{DSRL}~\citep{DSRL}: A latent steering method for diffusion- and flow-based policies that keeps the base policy frozen and applies RL in its initial latent-noise space rather than directly modifying the generated actions.\\
\noindent\textbf{PA-RL}~\citep{PA-RL}: A policy-agnostic RL finetuning method that uses a learned critic to improve action candidates through re-ranking and local optimization, and then distills the improved actions back into the base policy.\\
\noindent\textbf{OGPO}~\citep{ogpo}: An off-policy full-finetuning method for diffusion- and flow-based policies. It learns off-policy critics for data reuse and uses their values to optimize the full diffusion or flow generation process through a modified PPO objective.\\
\noindent\textbf{ResFiT}~\citep{ankile2025residual}: A demo-augmented residual RL method that keeps the base policy frozen, constrains residual magnitude for stable exploration, and jointly trains on demonstration and online data; we adapt its original per-step residual formulation to operate over action chunks. \\
\noindent\textbf{DICE-RL}~\citep{DICE-RL}:  Also a demo-augmented residual RL method that combines value-guided action selection with selective behavior regularization to reinforce high-value behaviors while remaining close to the pretrained policy.

\begin{figure*}[!t]
      \centering
      \includegraphics[width = 1.0\linewidth]{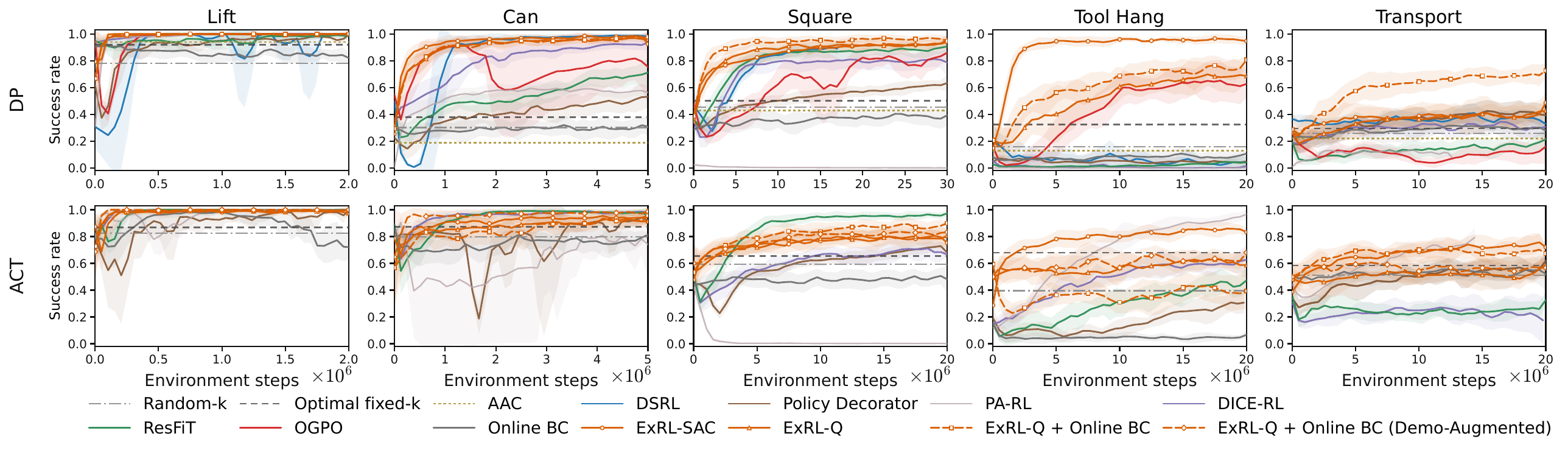}
      \caption{Online adaptation of DP and ACT on \texttt{Robomimic} with chunk size $H=32$.}
      \label{fig:robomimic_comparison_ta32}
\end{figure*}

\noindent\textbf{Simulation benchmarks and setup.}
We consider three simulation benchmarks: \texttt{Robomimic} \citep{mandlekar2021matters}, spanning five tabletop manipulation tasks from simple tasks to challenging long-horizon tasks such as \texttt{Tool Hang} and \texttt{Transport}; \texttt{Adroit} \citep{rajeswaran2017learning}, spanning four high-frequency dexterous hand manipulation tasks ($H=32$); and a \texttt{LIBERO} \citep{liu2023libero} task following the DSRL setup, except that we use $H=50$ instead of $H=10$. On \texttt{Robomimic}, the main experiments use $H=50$ for DP and ACT; Appendix Figure~\ref{fig:robomimic_comparison_ta32} reports additional $H=32$ results for both policies. Following prior work \citep{DSRL,ogpo}, we use base-policy checkpoints that leave meaningful headroom for RL improvement. Our goal on these benchmarks is not to maximize base-policy performance, but to use them as a controlled testbed for evaluating long-chunk policy adaptation. For each variant of ExRL, we keep most hyperparameters fixed across tasks and make only a few task-level adjustments when necessary, such as the discount factor $\gamma$ and network width. Appendix~\ref{sec:experimental_details} provides task details, policy architectures, baseline adaptations, and hyperparameter settings. For each run in the plot, we report the average success rate and 95\% confidence intervals over three seeds.

\subsection{Sample-Efficient Adaptation in Simulation}

Figures~\ref{fig:robomimic_comparison}, \ref{fig:adroit_comparison}, and~\ref{fig:libero_comparison} show that ExRL substantially improves frozen base policies, often reaching success rates close to $100\%$ by learning only the execution length. The gains over the optimal fixed-$k$ baseline are particularly large on DP \texttt{Can}, \texttt{Square}, and \texttt{Tool Hang}, and across all four \texttt{LIBERO} policies. As these improvements require no changes to the actions predicted by the base policy, they validate the importance and expressiveness of execution-length control as an adaptation interface.

\begin{table}[!t]
\caption{Real-world evaluation of $\pi_{0.5}$. Entries show successful trials / total trials for stacking, folding, and insertion, and completed stages / total stages for organization.}
\label{tab:real_robot_eval}
\centering
\footnotesize
\setlength{\tabcolsep}{2pt}
\renewcommand{\arraystretch}{1.12}
\begin{tabular*}{\linewidth}{@{\extracolsep{\fill}}lcccc@{}}
\toprule
 & Stack cube & Clothing folding & Power insertion & Parts organization \\
\midrule
$k=1$ & 0/20 & 0/20 & 7/25 & 0/125 \\
$k=10$ & 9/20 & 12/20 & 10/25 & 69/125 \\
$k=50$ & 1/20 & 11/20 & 0/25 & 46/125 \\
ExRL-Q & \textbf{17/20} & \textbf{17/20} & \textbf{22/25} & \textbf{104/125} \\
\bottomrule
\end{tabular*}

\end{table}

Compared with other RL methods, ExRL often learns substantially faster, particularly early in online adaptation. On \texttt{LIBERO}, for example, it improves $\pi_{0.5}$ from nearly $0\%$ to over $90\%$ success within $200{,}000$ environment steps. The advantage is also pronounced on the challenging \texttt{Tool Hang} task, where ExRL-SAC achieves high success with both DP and ACT; with DP, it exceeds $90\%$ success while all other baselines remain at around $20\%$ or below.

Online BC further improves both sample efficiency and final performance, particularly on DP \texttt{Transport} and several Adroit tasks. In these settings, ExRL-Q + Online BC substantially outperforms both frozen-base ExRL-Q and Online BC alone, supporting the value of distilling the improved behaviors discovered by ExRL back into the base policy.

With demonstration augmentation, ExRL-Q + Online BC rapidly reaches near-perfect success on ACT \texttt{Can} and all four Adroit tasks, competitive with the strongest demonstration-based baselines.

\begingroup
\setlength{\intextsep}{2pt}
\setlength{\abovecaptionskip}{3pt}
\setlength{\columnsep}{6pt}
\emergencystretch=1em
\rightskip=0pt plus 2em
\hyphenpenalty=50
\exhyphenpenalty=50
\begin{wrapfigure}[8]{r}{0.40\columnwidth}
\vspace{-4pt}
\centering
\includegraphics[width=\linewidth,trim=0 0.9bp 0 0,clip]{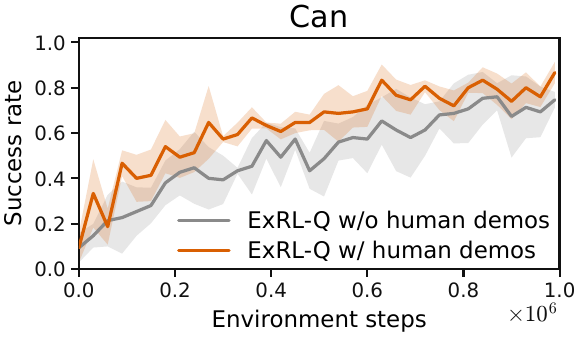}
\caption{Human demonstrations accelerate ExRL-Q on \texttt{Can} ($H=50$).}
\label{fig:offline_human_demos}
\setcounter{figure}{9}
\end{wrapfigure}
Figure~\ref{fig:offline_human_demos} further isolates the benefit of demonstrations for execution-length learning: adding human experience to RL replay accelerates early adaptation on \texttt{Can} while keeping the base policy frozen.

The \texttt{LIBERO} results also highlight the generality of ExRL. The same execution-length formulation enables efficient adaptation of flow-based VLAs, autoregressive VLAs, and a world action model, despite their substantially different architectures and action-generation processes.

\endgroup

\subsection{Sample-Efficient Real-World Adaptation}

We further adapt $\pi_{0.5}$ with chunk size $H=50$ on four real-world tasks (Figure~\ref{fig:real_world_adaptation}, right; Table~\ref{tab:real_robot_eval}). \textit{Power insertion} requires contact-rich manipulation, while \textit{clothing folding} involves manipulating deformable objects. Both clothing folding and \textit{parts organization} require long-horizon coordination. We report success rate for stacking, folding, and insertion, and the fraction of completed stages for organization. Further setup details are provided in Appendix~\ref{sec:real_world_details}.

The sample efficiency of ExRL carries over to real-world adaptation. With the base policy frozen, ExRL-Q approaches $100\%$ success on stack cube with only ten minutes of physical interaction, and reaches nearly $90\%$ on the long-horizon clothing folding task within $85$ minutes. Power insertion and parts organization similarly improve within a few hundred rollouts. Final evaluations show substantial gains over fixed-length execution across all four tasks (Table~\ref{tab:real_robot_eval}).

\subsection{Ablation Studies}
\label{sec:ablations}

Accounting for the selected execution length in reward accumulation and discounting improves learning over a TD target that treats each decision as a single step (Figure~\ref{fig:other_ablations}, leftmost panel). The predicted chunk is also essential to the decision: conditioning on both the observation and the chunk raises success from around $30\%$ to nearly $90\%$ compared with observations alone (second panel). Thus, the critic benefits from knowing what the base policy proposes, beyond the current state. Allowing $k=0$ offers little additional benefit over positive execution lengths (third panel), while similar learning curves across network widths from $128$ to $2048$ suggest that ExRL does not require precise tuning of network capacity (rightmost panel).

\subsection{What Execution Length Patterns Does ExRL Learn?}
\label{sec:execution_patterns}

ExRL improves a policy without changing its predicted actions, raising the question of how execution-length decisions reshape its behavior. Figure~\ref{fig:visulization}(a) reveals a consistent pattern: across \texttt{Lift}, \texttt{Can}, \texttt{Square}, and real-world insertion, ExRL selects shorter executions near the object or target than during free motion. The mean length roughly halves in simulation and falls to about a quarter on real-world insertion. ExRL therefore concentrates replanning near interaction, while allowing longer motions elsewhere.

Short executions, however, need not serve the same purpose. Two \texttt{Square} rollouts illustrate how ExRL decides when to continue with a longer chunk. In Figure~\ref{fig:visulization}(b), it repeatedly truncates unsuitable predictions with little physical motion, then accepts a chunk that predicts a useful grasp-and-lift motion. Here, frequent replanning provides new candidate motions from nearly the same state. In Figure~\ref{fig:visulization}(c), short executions instead change the state through small corrections, bringing the nut closer to the peg before ExRL switches to longer execution. The distinction is between waiting for a useful prediction and moving into a state from which a longer prediction becomes useful. Both behaviors depend on the current state and the proposed chunk, rather than a fixed preference for short execution.

Figure~\ref{fig:real_execution_patterns} in the appendix illustrates how this balance unfolds across four real-world tasks. Further analysis shows that shuffling the same execution lengths across states and chunks reduces success (Appendix~\ref{app:conditional_assignment}). It also suggests a complementary reason to preserve longer execution: under \emph{feature aliasing}, locally similar policy features can repeatedly produce similar first actions, so frequent replanning may interrupt useful progression through a chunk (Appendix~\ref{app:feature_aliasing}).

\section{Conclusion}
We introduced ExRL, which shifts online adaptation of long-chunk policies from action generation to execution-length control. Conditioning this scalar decision on the state and predicted chunk provides substantial behavioral flexibility, while keeping optimization low-dimensional as chunk size grows. Across simulation and real-world tasks, ExRL enables sample-efficient improvement of diverse frozen policies, often surpassing their best fixed execution length. Online imitation learning and offline demonstrations further extend these gains and support continued base-policy improvement during RL. These findings show that learning how to execute a policy's predictions can unlock capabilities already present in the policy, making execution length a practical interface for adapting long-chunk robot policies with limited online interaction.

\section{Limitations}
Our real-world experiments keep the base policy frozen and do not yet incorporate online BC. Performance therefore remains constrained by the actions the base policy can propose, and several tasks still fall short of $100\%$ success. Extending online BC to these experiments may help address the remaining failures. We also adapt each task separately and have not evaluated joint multi-task adaptation or generalization to unseen tasks. Recent work demonstrates that sufficiently diverse multi-task experience and value-based RL can improve performance on unseen scenes and objects in simulation~\citep{korniak2026reinforcementlearningenablesscaling}; whether execution-length adaptation offers similar benefits remains open. Finally, we have not tested ExRL together with asynchronous inference methods such as real-time chunking (RTC)~\citep{RTC}. Understanding how adaptive replanning interacts with overlapping inference and execution is an important direction for extending ExRL to latency-sensitive deployments.

\bibliography{iclr2027_conference}

@article{pi_0,
  title={{$\pi_0$}: A Vision-Language-Action Flow Model for General Robot Control},
  author={Black, Kevin and Brown, Noah and Driess, Danny and Esmail, Adnan and Equi, Michael and Finn, Chelsea and Fusai, Niccolo and Groom, Lachy and Hausman, Karol and Ichter, Brian and others},
  journal={arXiv preprint arXiv:2410.24164},
  year={2024}
}

@article{pi0.5,
  title={{$\pi_{0.5}$}: a Vision-Language-Action Model with Open-World Generalization},
  author={Black, Kevin and Brown, Noah and Darpinian, James and Dhabalia, Karan and Driess, Danny and Esmail, Adnan and Equi, Michael and Finn, Chelsea and Fusai, Niccolo and others},
  journal={arXiv preprint arXiv:2504.16054},
  year={2025}
}

@article{pi0.7,
  title = {{$\pi_{0.7}$}: A Steerable Generalist Robotic Foundation Model with Emergent Capabilities},
  author={Ai, Bo and Amin, Ali and Aniceto, Raichelle and Balakrishna, Ashwin and Balke, Greg and Black, Kevin and Bokinsky, George and Cao, Shihao and Charbonnier, Thomas and others},
  journal={arXiv preprint arXiv:2604.15483},
  year={2026}
}

@article{gr00t,
  title={{GR00T N1}: An open foundation model for generalist humanoid robots},
  author={Bjorck, Johan and Casta{\~n}eda, Fernando and Cherniadev, Nikita and Da, Xingye and Ding, Runyu and Fan, Linxi and Fang, Yu and Fox, Dieter and Hu, Fengyuan and Huang, Spencer and others},
  journal={arXiv preprint arXiv:2503.14734},
  year={2025}
}

@article{simchowitz2025pitfalls,
  title={The pitfalls of imitation learning when actions are continuous},
  author={Simchowitz, Max and Pfrommer, Daniel and Jadbabaie, Ali},
  journal={arXiv preprint arXiv:2503.09722},
  year={2025}
}

@article{pi0.6,
  title={{$\pi^*_{0.6}$}: a {VLA} That Learns From Experience},
  author={Amin, Ali and Aniceto, Raichelle and Balakrishna, Ashwin and Black, Kevin and Conley, Ken and Connors, Grace and Darpinian, James and Dhabalia, Karan and DiCarlo, Jared and others},
  journal={arXiv preprint arXiv:2511.14759},
  year={2025}
}

@article{openvla,
  title={{OpenVLA}: An open-source vision-language-action model},
  author={Kim, Moo Jin and Pertsch, Karl and Karamcheti, Siddharth and Xiao, Ted and Balakrishna, Ashwin and Nair, Suraj and Rafailov, Rafael and Foster, Ethan and Lam, Grace and Sanketi, Pannag and others},
  journal={arXiv preprint arXiv:2406.09246},
  year={2024}
}

@article{OpenAI-o1,
  title={{OpenAI o1} system card},
  author={Jaech, Aaron and Kalai, Adam and Lerer, Adam and Richardson, Adam and El-Kishky, Ahmed and Low, Aiden and Helyar, Alec and Madry, Aleksander and Beutel, Alex and Carney, Alex and others},
  journal={arXiv preprint arXiv:2412.16720},
  year={2024}
}

@article{DeepSeek-r1,
  title={{DeepSeek-R1}: Incentivizing reasoning capability in {LLMs} via reinforcement learning},
  author={Guo, Daya and Yang, Dejian and Zhang, Haowei and Song, Junxiao and Wang, Peiyi and Zhu, Qihao and Xu, Runxin and Zhang, Ruoyu and Ma, Shirong and Bi, Xiao and others},
  journal={arXiv preprint arXiv:2501.12948},
  year={2025}
}

@article{ACT,
  title={Learning fine-grained bimanual manipulation with low-cost hardware},
  author={Zhao, Tony Z and Kumar, Vikash and Levine, Sergey and Finn, Chelsea},
  journal={arXiv preprint arXiv:2304.13705},
  year={2023}
}

@article{DP,
  title={Diffusion policy: Visuomotor policy learning via action diffusion},
  author={Chi, Cheng and Xu, Zhenjia and Feng, Siyuan and Cousineau, Eric and Du, Yilun and Burchfiel, Benjamin and Tedrake, Russ and Song, Shuran},
  journal={The International Journal of Robotics Research},
  volume={44},
  number={10-11},
  pages={1684--1704},
  year={2025},
  publisher={Sage Publications Sage UK: London, England}
}

@article{PA-RL,
  title={Policy agnostic {RL}: Offline {RL} and online {RL} fine-tuning of any class and backbone},
  author={Mark, Max Sobol and Gao, Tian and Sampaio, Georgia Gabriela and Srirama, Mohan Kumar and Sharma, Archit and Finn, Chelsea and Kumar, Aviral},
  journal={arXiv preprint arXiv:2412.06685},
  year={2024}
}

@article{fast,
  title={{FAST}: Efficient action tokenization for vision-language-action models},
  author={Pertsch, Karl and Stachowicz, Kyle and Ichter, Brian and Driess, Danny and Nair, Suraj and Vuong, Quan and Mees, Oier and Finn, Chelsea and Levine, Sergey},
  journal={arXiv preprint arXiv:2501.09747},
  year={2025}
}

@article{ouyang2022training,
  title={Training language models to follow instructions with human feedback},
  author={Ouyang, Long and Wu, Jeffrey and Jiang, Xu and Almeida, Diogo and Wainwright, Carroll and Mishkin, Pamela and Zhang, Chong and Agarwal, Sandhini and Slama, Katarina and Ray, Alex and others},
  journal={Advances in neural information processing systems},
  volume={35},
  pages={27730--27744},
  year={2022}
}

@article{DSRL,
  title={Steering your diffusion policy with latent space reinforcement learning},
  author={Wagenmaker, Andrew and Nakamoto, Mitsuhiko and Zhang, Yunchu and Park, Seohong and Yagoub, Waleed and Nagabandi, Anusha and Gupta, Abhishek and Levine, Sergey},
  journal={arXiv preprint arXiv:2506.15799},
  year={2025}
}

@article{cosmos,
  title={Cosmos policy: Fine-tuning video models for visuomotor control and planning},
  author={Kim, Moo Jin and Gao, Yihuai and Lin, Tsung-Yi and Lin, Yen-Chen and Ge, Yunhao and Lam, Grace and Liang, Percy and Song, Shuran and Liu, Ming-Yu and Finn, Chelsea and others},
  journal={arXiv preprint arXiv:2601.16163},
  year={2026}
}

@inproceedings{liu2025bidirectional,
  title={Bidirectional decoding: Improving action chunking via guided test-time sampling},
  author={Liu, Yuejiang and Hamid, Jubayer and Xie, Annie and Lee, Yoonho and Du, Max and Finn, Chelsea},
  booktitle={International Conference on Learning Representations},
  volume={2025},
  pages={4594--4627},
  year={2025}
}

@article{zhang2025action,
  title={Action Chunking and Exploratory Data Collection Yield Exponential Improvements in Behavior Cloning for Continuous Control},
  author={Zhang, Thomas T and Pfrommer, Daniel and Pan, Chaoyi and Matni, Nikolai and Simchowitz, Max},
  journal={arXiv preprint arXiv:2507.09061},
  year={2025}
}

@article{watkins1992q,
  title={{Q}-learning},
  author={Watkins, Christopher JCH and Dayan, Peter},
  journal={Machine learning},
  volume={8},
  number={3--4},
  pages={279--292},
  year={1992},
  publisher={Springer}
}

@article{christodoulou2019soft,
  title={Soft actor-critic for discrete action settings},
  author={Christodoulou, Petros},
  journal={arXiv preprint arXiv:1910.07207},
  year={2019}
}

@article{mnih2013playing,
  title={Playing {Atari} with deep reinforcement learning},
  author={Mnih, Volodymyr and Kavukcuoglu, Koray and Silver, David and Graves, Alex and Antonoglou, Ioannis and Wierstra, Daan and Riedmiller, Martin},
  journal={arXiv preprint arXiv:1312.5602},
  year={2013}
}

@article{ZPRL,
  title={Beyond Action Residuals: Real-World Robot Policy Steering via Bottleneck Latent Reinforcement Learning},
  author={Yu, Dongjie and Lei, Kun and Jiang, Zhennan and Pan, Jia and Xu, Huazhe},
  journal={arXiv preprint arXiv:2605.19919},
  year={2026}
}

@article{RTC,
  title={Real-time execution of action chunking flow policies},
  author={Black, Kevin and Galliker, Manuel and Levine, Sergey},
  journal={Advances in Neural Information Processing Systems},
  volume={38},
  pages={33383--33407},
  year={2025}
}

@inproceedings{ross2011reduction,
  title={A reduction of imitation learning and structured prediction to no-regret online learning},
  author={Ross, St{\'e}phane and Gordon, Geoffrey and Bagnell, Drew},
  booktitle={Proceedings of the fourteenth international conference on artificial intelligence and statistics},
  pages={627--635},
  year={2011},
  organization={JMLR Workshop and Conference Proceedings}
}

@inproceedings{zhang2018deep,
  title={Deep imitation learning for complex manipulation tasks from virtual reality teleoperation},
  author={Zhang, Tianhao and McCarthy, Zoe and Jow, Owen and Lee, Dennis and Chen, Xi and Goldberg, Ken and Abbeel, Pieter},
  booktitle={2018 IEEE international conference on robotics and automation (ICRA)},
  pages={5628--5635},
  year={2018},
  organization={Ieee}
}

@inproceedings{atkeson1997robot,
  title={Robot learning from demonstration},
  author={Atkeson, Christopher G and Schaal, Stefan},
  booktitle={ICML},
  volume={97},
  pages={12--20},
  year={1997}
}

@inproceedings{florence2022implicit,
  title={Implicit behavioral cloning},
  author={Florence, Pete and Lynch, Corey and Zeng, Andy and Ramirez, Oscar A and Wahid, Ayzaan and Downs, Laura and Wong, Adrian and Lee, Johnny and Mordatch, Igor and Tompson, Jonathan},
  booktitle={Conference on robot learning},
  pages={158--168},
  year={2022},
  organization={PMLR}
}

@article{shafiullah2022behavior,
  title={Behavior transformers: Cloning $ k $ modes with one stone},
  author={Shafiullah, Nur Muhammad and Cui, Zichen and Altanzaya, Ariuntuya Arty and Pinto, Lerrel},
  journal={Advances in neural information processing systems},
  volume={35},
  pages={22955--22968},
  year={2022}
}

@article{rt-1,
  title={{RT-1}: Robotics transformer for real-world control at scale},
  author={Brohan, Anthony and Brown, Noah and Carbajal, Justice and Chebotar, Yevgen and Dabis, Joseph and Finn, Chelsea and Gopalakrishnan, Keerthana and Hausman, Karol and Herzog, Alex and Hsu, Jasmine and others},
  journal={arXiv preprint arXiv:2212.06817},
  year={2022}
}

@inproceedings{oxe,
  title={{Open X-Embodiment}: Robotic learning datasets and {RT-X} models},
  author={O’Neill, Abby and Rehman, Abdul and Maddukuri, Abhiram and Gupta, Abhishek and Padalkar, Abhishek and Lee, Abraham and Pooley, Acorn and Gupta, Agrim and Mandlekar, Ajay and Jain, Ajinkya and others},
  booktitle={2024 IEEE International Conference on Robotics and Automation (ICRA)},
  pages={6892--6903},
  year={2024},
  organization={IEEE}
}

@inproceedings{bridge,
  title={{BridgeData V2}: A dataset for robot learning at scale},
  author={Walke, Homer Rich and Black, Kevin and Zhao, Tony Z and Vuong, Quan and Zheng, Chongyi and Hansen-Estruch, Philippe and He, Andre Wang and Myers, Vivek and Kim, Moo Jin and Du, Max and others},
  booktitle={Conference on Robot Learning},
  pages={1723--1736},
  year={2023},
  organization={PMLR}
}

@article{droid,
  title={{DROID}: A large-scale in-the-wild robot manipulation dataset},
  author={Khazatsky, Alexander and Pertsch, Karl and Nair, Suraj and Balakrishna, Ashwin and Dasari, Sudeep and Karamcheti, Siddharth and Nasiriany, Soroush and Srirama, Mohan Kumar and Chen, Lawrence Yunliang and Ellis, Kirsty and others},
  journal={arXiv preprint arXiv:2403.12945},
  year={2024}
}

@article{octo,
  title={{Octo}: An open-source generalist robot policy},
  author={{Octo Model Team} and Ghosh, Dibya and Walke, Homer and Pertsch, Karl and Black, Kevin and Mees, Oier and Dasari, Sudeep and Hejna, Joey and Kreiman, Tobias and Xu, Charles and others},
  journal={arXiv preprint arXiv:2405.12213},
  year={2024}
}

@article{mandlekar2021matters,
  title={What matters in learning from offline human demonstrations for robot manipulation},
  author={Mandlekar, Ajay and Xu, Danfei and Wong, Josiah and Nasiriany, Soroush and Wang, Chen and Kulkarni, Rohun and Fei-Fei, Li and Savarese, Silvio and Zhu, Yuke and Mart{\'\i}n-Mart{\'\i}n, Roberto},
  journal={arXiv preprint arXiv:2108.03298},
  year={2021}
}

@article{khansari2011learning,
  title={Learning stable nonlinear dynamical systems with {Gaussian} mixture models},
  author={Khansari-Zadeh, S Mohammad and Billard, Aude},
  journal={IEEE Transactions on Robotics},
  volume={27},
  number={5},
  pages={943--957},
  year={2011},
  publisher={IEEE}
}

@inproceedings{rt-2,
  title={{RT-2}: Vision-language-action models transfer web knowledge to robotic control},
  author={Zitkovich, Brianna and Yu, Tianhe and Xu, Sichun and Xu, Peng and Xiao, Ted and Xia, Fei and Wu, Jialin and Wohlhart, Paul and Welker, Stefan and Wahid, Ayzaan and others},
  booktitle={Conference on Robot Learning},
  pages={2165--2183},
  year={2023},
  organization={PMLR}
}

@article{smolvla,
  title={{SmolVLA}: A vision-language-action model for affordable and efficient robotics},
  author={Shukor, Mustafa and Aubakirova, Dana and Capuano, Francesco and Kooijmans, Pepijn and Palma, Steven and Zouitine, Adil and Aractingi, Michel and Pascal, Caroline and Russi, Martino and Marafioti, Andres and others},
  journal={arXiv preprint arXiv:2506.01844},
  year={2025}
}

@article{AAC,
  title={Adaptive action chunking at inference-time for vision-language-action models},
  author={Liang, Yuanchang and Wang, Xiaobo and Wang, Kai and Wang, Shuo and Peng, Xiaojiang and Chen, Haoyu and Chua, David Kim Huat and Vadakkepat, Prahlad},
  journal={arXiv preprint arXiv:2604.04161},
  year={2026}
}

@article{wang2026trust,
  title={When to Trust Imagination: Adaptive Action Execution for World Action Models},
  author={Wang, Rui and Zhang, Yue and Lin, Jiehong and Luo, Kuncheng and Wang, Jianan and Wang, Zhongrui and Qi, Xiaojuan},
  journal={arXiv preprint arXiv:2605.06222},
  year={2026}
}

@article{ACSAC,
  title={{ACSAC}: Adaptive Chunk Size Actor-Critic with Causal Transformer {Q}-Network},
  author={Chen, Qian and Zhao, Junqiao and Zhou, Hongtu and Yu, Hang and Zhao, Yanping and Ye, Chen and Chen, Guang},
  journal={arXiv preprint arXiv:2605.11009},
  year={2026}
}

@article{shin2026adaptive,
  title={Adaptive Action Chunking via Multi-Chunk {Q} Value Estimation},
  author={Shin, Yongjae and Chae, Jongseong and Kim, Seongmin and Park, Jongeui and Sung, Youngchul},
  journal={arXiv preprint arXiv:2605.10044},
  year={2026}
}

@article{li2025simplevla,
  title={{SimpleVLA-RL}: Scaling {VLA} training via reinforcement learning},
  author={Li, Haozhan and Zuo, Yuxin and Yu, Jiale and Zhang, Yuhao and Yang, Zhaohui and Zhang, Kaiyan and Zhu, Xuekai and Zhang, Yuchen and Chen, Tianxing and Cui, Ganqu and others},
  journal={arXiv preprint arXiv:2509.09674},
  year={2025}
}

@article{lu2025vla,
  title={{VLA-RL}: Towards masterful and general robotic manipulation with scalable reinforcement learning},
  author={Lu, Guanxing and Guo, Wenkai and Zhang, Chubin and Zhou, Yuheng and Jiang, Haonan and Gao, Zifeng and Tang, Yansong and Wang, Ziwei},
  journal={arXiv preprint arXiv:2505.18719},
  year={2025}
}

@article{chen2025pirl,
  title={{$\pi_{\texttt{RL}}$}: Online {RL} fine-tuning for flow-based vision-language-action models},
  author={Chen, Kang and Liu, Zhihao and Zhang, Tonghe and Guo, Zhen and Xu, Si and Lin, Hao and Zang, Hongzhi and Zhang, Quanlu and Yu, Zhaofei and Fan, Guoliang and others},
  journal={arXiv preprint arXiv:2510.25889},
  year={2025}
}

@inproceedings{SAC,
  title={Soft actor-critic: Off-policy maximum entropy deep reinforcement learning with a stochastic actor},
  author={Haarnoja, Tuomas and Zhou, Aurick and Abbeel, Pieter and Levine, Sergey},
  booktitle={International conference on machine learning},
  pages={1861--1870},
  year={2018},
  organization={Pmlr}
}

@inproceedings{TD3,
  title={Addressing function approximation error in actor-critic methods},
  author={Fujimoto, Scott and van Hoof, Herke and Meger, David},
  booktitle={International conference on machine learning},
  pages={1587--1596},
  year={2018},
  organization={PMLR}
}

@inproceedings{dppo,
  title={Diffusion policy policy optimization},
  author={Ren, Allen and Lidard, Justin and Ankile, Lars and Simeonov, Anthony and Agrawal, Pulkit and Majumdar, Anirudha and Burchfiel, Benjamin and Dai, Hongkai and Simchowitz, Max},
  booktitle={International Conference on Learning Representations},
  volume={2025},
  pages={77288--77329},
  year={2025}
}

@article{expo,
  title={{EXPO}: Stable reinforcement learning with expressive policies},
  author={Dong, Perry and Li, Qiyang and Sadigh, Dorsa and Finn, Chelsea},
  journal={arXiv preprint arXiv:2507.07986},
  year={2025}
}

@article{rl-token,
  title={{RL Token}: Bootstrapping Online {RL} with Vision-Language-Action Models},
  author={Xu, Charles and Springenberg, Jost Tobias and Equi, Michael and Amin, Ali and Esmail, Adnan and Levine, Sergey and Ke, Liyiming},
  journal={arXiv preprint arXiv:2604.23073},
  year={2026}
}

@inproceedings{ankile2025imitation,
  title={From imitation to refinement---residual {RL} for precise assembly},
  author={Ankile, Lars and Simeonov, Anthony and Shenfeld, Idan and Torne, Marcel and Agrawal, Pulkit},
  booktitle={2025 IEEE International Conference on Robotics and Automation (ICRA)},
  pages={01--08},
  year={2025},
  organization={IEEE}
}

@article{ankile2025residual,
  title={Residual off-policy {RL} for finetuning behavior cloning policies},
  author={Ankile, Lars and Jiang, Zhenyu and Duan, Rocky and Shi, Guanya and Abbeel, Pieter and Nagabandi, Anusha},
  journal={arXiv preprint arXiv:2509.19301},
  year={2025}
}

@article{ogpo,
  title={{OGPO}: Sample Efficient Full-Finetuning of Generative Control Policies},
  author={Patil, Sarvesh and Nakamoto, Mitsuhiko and Agarwal, Manan and Saxena, Shashwat and Zhang, Jesse and Anantharaman, Giri and Winston, Cleah and Pan, Chaoyi and Chen, Douglas and Huang, Nai-Chieh and others},
  journal={arXiv preprint arXiv:2605.03065},
  year={2026}
}

@article{rl-100,
  title={{RL-100}: Performant robotic manipulation with real-world reinforcement learning},
  author={Lei, Kun and Li, Huanyu and Yu, Dongjie and Wei, Zhenyu and Guo, Lingxiao and Jiang, Zhennan and Wang, Ziyu and Liang, Shiyu and Xu, Huazhe},
  journal={arXiv preprint arXiv:2510.14830},
  year={2025}
}

@article{dong2025matters,
  title={What Matters for Batch Online Reinforcement Learning in Robotics?},
  author={Dong, Perry and Mirchandani, Suvir and Sadigh, Dorsa and Finn, Chelsea},
  journal={arXiv preprint arXiv:2505.08078},
  year={2025}
}

@article{SMDP,
  title={Between {MDPs} and semi-{MDPs}: A framework for temporal abstraction in reinforcement learning},
  author={Sutton, Richard S and Precup, Doina and Singh, Satinder},
  journal={Artificial intelligence},
  volume={112},
  number={1-2},
  pages={181--211},
  year={1999},
  publisher={Elsevier}
}

@article{li2026reinforcement,
  title={Reinforcement learning with action chunking},
  author={Li, Qiyang and Zhou, Zhiyuan Paul and Levine, Sergey},
  journal={Advances in Neural Information Processing Systems},
  volume={38},
  pages={55518--55553},
  year={2025}
}

@article{CQL,
  title={Conservative {Q}-learning for offline reinforcement learning},
  author={Kumar, Aviral and Zhou, Aurick and Tucker, George and Levine, Sergey},
  journal={Advances in neural information processing systems},
  volume={33},
  pages={1179--1191},
  year={2020}
}

@article{ho2020denoising,
  title={Denoising diffusion probabilistic models},
  author={Ho, Jonathan and Jain, Ajay and Abbeel, Pieter},
  journal={Advances in neural information processing systems},
  volume={33},
  pages={6840--6851},
  year={2020}
}

@article{lipman2022flow,
  title={Flow matching for generative modeling},
  author={Lipman, Yaron and Chen, Ricky TQ and Ben-Hamu, Heli and Nickel, Maximilian and Le, Matt},
  journal={arXiv preprint arXiv:2210.02747},
  year={2022}
}

@article{Policy-decorator,
  title={Policy decorator: Model-agnostic online refinement for large policy model},
  author={Yuan, Xiu and Mu, Tongzhou and Tao, Stone and Fang, Yunhao and Zhang, Mengke and Su, Hao},
  journal={arXiv preprint arXiv:2412.13630},
  year={2024}
}

@article{dreamzero,
  title={World action models are zero-shot policies},
  author={Ye, Seonghyeon and Ge, Yunhao and Zheng, Kaiyuan and Gao, Shenyuan and Yu, Sihyun and Kurian, George and Indupuru, Suneel and Tan, You Liang and Zhu, Chuning and Xiang, Jiannan and others},
  journal={arXiv preprint arXiv:2602.15922},
  year={2026}
}

@article{lazzati2026does,
  title={Why Does Action Chunking Improve Behavioral Cloning Performance in Robotic Control?},
  author={Lazzati, Filippo and Stachowicz, Kyle and Chen, William and Metelli, Alberto Maria and Wagenmaker, Andrew and Levine, Sergey},
  journal={arXiv preprint arXiv:2608.02547},
  year={2026}
}

@article{sunday2026act2preview,
  author  = {{Sunday Robotics}},
  title   = {{ACT-2} Preview: Generalizing Reliability},
  journal = {Sunday Robotics Blog},
  year    = {2026},
  month   = {jul},
  url     = {https://www.sunday.ai/blog/act-2-preview}
}

@article{DICE-RL,
  title={From prior to pro: Efficient skill mastery via distribution contractive {RL} finetuning},
  author={Sun, Zhanyi and Song, Shuran},
  journal={arXiv preprint arXiv:2603.10263},
  year={2026}
}

@article{zhao2026dynamic,
  title={Dynamic Execution Horizon Prediction for Chunk-based Robot Policies},
  author={Zhao, Yuchi and Bogdanovic, Miroslav and Sohal, Arjun and Tao, Liyu and Darvish, Kourosh and Aspuru-Guzik, Al{\'a}n and Shkurti, Florian and Garg, Animesh},
  journal={arXiv preprint arXiv:2606.11408},
  year={2026}
}

@inproceedings{lakshminarayanan2017dynamic,
  title={Dynamic action repetition for deep reinforcement learning},
  author={Lakshminarayanan, Aravind and Sharma, Sahil and Ravindran, Balaraman},
  booktitle={Proceedings of the AAAI conference on artificial intelligence},
  volume={31},
  number={1},
  year={2017}
}

@inproceedings{biedenkapp2021temporl,
  title={{TempoRL}: Learning when to act},
  author={Biedenkapp, Andr{\'e} and Rajan, Raghu and Hutter, Frank and Lindauer, Marius},
  booktitle={International Conference on Machine Learning},
  pages={914--924},
  year={2021},
  organization={PMLR}
}

@article{weng2025temporal,
  title={Temporal action selection for action chunking},
  author={Weng, Yueyang and Zhang, Xiaopeng and Mu, Yongjin and Zhu, Yingcong and Li, Yanjie},
  journal={arXiv preprint arXiv:2511.04421},
  year={2025}
}

@article{rajeswaran2017learning,
  title={Learning complex dexterous manipulation with deep reinforcement learning and demonstrations},
  author={Rajeswaran, Aravind and Kumar, Vikash and Gupta, Abhishek and Vezzani, Giulia and Schulman, John and Todorov, Emanuel and Levine, Sergey},
  journal={arXiv preprint arXiv:1709.10087},
  year={2017}
}

@article{liu2023libero,
  title={{LIBERO}: Benchmarking knowledge transfer for lifelong robot learning},
  author={Liu, Bo and Zhu, Yifeng and Gao, Chongkai and Feng, Yihao and Liu, Qiang and Zhu, Yuke and Stone, Peter},
  journal={Advances in Neural Information Processing Systems},
  volume={36},
  pages={44776--44791},
  year={2023}
}

@article{fu2020d4rl,
  title={{D4RL}: Datasets for Deep Data-Driven Reinforcement Learning},
  author={Fu, Justin and Kumar, Aviral and Nachum, Ofir and Tucker, George and Levine, Sergey},
  journal={arXiv preprint arXiv:2004.07219},
  year={2020}
}

@misc{korniak2026reinforcementlearningenablesscaling,
  title = {Reinforcement Learning Enables Scaling Generalist Robot Policy Improvement},
  author = {Michal Korniak and
            Marco Bagatella and
            Ralf R{\"o}mer and
            Michal Nauman and
            Yarden As and
            Pranav Vignesh and
            Manan Agarwal and
            Leander Diaz-Bone and
            Piotr Zalewski and
            Miko{\l}aj Czarnecki and
            Lenart Treven and
            Shashwat Saxena and
            Sreyas Venkataraman and
            Sarvesh Patil and
            Zhiyuan Zhou and
            Daniel Marta and
            Oier Mees and
            Nur Muhammad Mahi Shafiullah and
            Carmelo Sferrazza and
            Bhavya Sukhija and
            Angela P. Schoellig and
            Andreas Krause and
            Max Simchowitz and
            Andrew Wagenmaker},
  year = {2026},
  note = {arXiv:2611.06905},
  eprint = {2611.06905},
  archivePrefix = {arXiv},
  primaryClass = {cs.LG},
  url = {https://arxiv.org/abs/2611.06905}
}
\bibliographystyle{plainnat}

\appendices
\setcounter{dbltopnumber}{4}
\renewcommand{\dbltopfraction}{0.8}

\begin{algorithm}[!htbp]
\caption{Execution-Length Control via Q-Learning (ExRL-Q)}
\label{alg:exrl_q_full}
\begin{algorithmic}[1]
\State \textbf{input:} pretrained action-chunked policy $\pi_{\mathrm{base}}$, online environment $\mathcal{M}$, execution-length space $\mathcal{K}=\{1,\ldots,H\}$
\State Initialize replay buffer $\mathcal{D}_{\mathrm{replay}}$
\State Initialize execution-length critic $Q_\phi$
\State Initialize target critic $Q_{\bar{\phi}}$ with $\bar{\phi}\leftarrow\phi$
\For{each training iteration}
    \LComment{Online Rollout}
    \State Sample an action chunk $a_{t:t+H}\sim\pi_{\mathrm{base}}(\cdot\mid s_t)$ at the current query state $s_t$
    \State Select execution length $k_t$ with Equation (6)
    \State Execute the first $k_t$ actions in the chunk and observe
    \Statex \hspace{1.5em} $\hat r_t \leftarrow \sum_{i=0}^{k_t-1}\gamma^i r(s_{t+i},a_{t+i})$ and $s' \leftarrow s_{t+k_t}$
    \State Add $(s_t,a_{t:t+H},k_t,\hat r_t,s')$ to $\mathcal{D}_{\mathrm{replay}}$
    \LComment{Train Critic}
    \State Sample $(s,a,k,\hat r,s')\sim\mathcal{D}_{\mathrm{replay}}$ and sample $a'\sim\pi_{\mathrm{base}}(\cdot\mid s')$
    \State Compute next-state value $V(s',a')$ with Equation (4)
    \State Compute TD target $y_t$ with Equation~(\ref{eq:duration_td_target})
    \State Compute $\mathcal{L}_{\mathrm{critic}}(\phi)$ with CQL using Equation~(\ref{eq:critic_cql})
    \State Train critic $Q_\phi$ by minimizing $\mathcal{L}_{\mathrm{critic}}(\phi)$
    \State Periodically update target: $\bar{\phi} \leftarrow \tau\phi + (1-\tau)\bar{\phi}$
\EndFor
\State \textbf{return} $Q_\phi$
\end{algorithmic}
\end{algorithm}

\begin{algorithm}[!htbp]
\caption{Execution-Length Control via Discrete Soft Actor-Critic (ExRL-SAC)}
\label{alg:exrl_sac_full}
\begin{algorithmic}[1]
\State \textbf{input:} pretrained action-chunked policy $\pi_{\mathrm{base}}$, online environment $\mathcal{M}$, execution-length space $\mathcal{K}=\{1,\ldots,H\}$, temperature $\alpha$
\State Initialize replay buffer $\mathcal{D}_{\mathrm{replay}}$
\State Initialize execution-length critic $Q_\phi$
\State Initialize target critic $Q_{\bar{\phi}}$ with $\bar{\phi}\leftarrow\phi$
\State Initialize execution-length actor $\pi_\psi$
\For{each training iteration}
    \LComment{Online Rollout}
    \State Sample an action chunk $a_{t:t+H}\sim\pi_{\mathrm{base}}(\cdot\mid s_t)$ at the current query state $s_t$
    \State Sample execution length $k_t \sim \pi_\psi(\cdot\mid s_t,a_{t:t+H})$
    \State Execute the first $k_t$ actions in the chunk and observe
    \Statex \hspace{1.5em} $\hat r_t \leftarrow \sum_{i=0}^{k_t-1}\gamma^i r(s_{t+i},a_{t+i})$ and $s' \leftarrow s_{t+k_t}$
    \State Add $(s_t,a_{t:t+H},k_t,\hat r_t,s')$ to $\mathcal{D}_{\mathrm{replay}}$
    \LComment{Train Critic and Actor}
    \State Sample $(s,a,k,\hat r,s')\sim\mathcal{D}_{\mathrm{replay}}$ and sample $a'\sim\pi_{\mathrm{base}}(\cdot\mid s')$
    \State Compute the soft value $V(s',a')$ with Equation (7)
    \State Compute TD target $y_t$ with Equation~(\ref{eq:duration_td_target})
    \State Compute $\mathcal{L}_{\mathrm{critic}}(\phi)$ with CQL using Equation~(\ref{eq:critic_cql})
    \State Compute actor loss $\mathcal{L}_\pi(\psi)$ with Equation (8)
    \State Train critic $Q_\phi$ by minimizing $\mathcal{L}_{\mathrm{critic}}(\phi)$
    \State Train actor $\pi_\psi$ by minimizing $\mathcal{L}_\pi(\psi)$
    \State Periodically update target: $\bar{\phi} \leftarrow \tau\phi + (1-\tau)\bar{\phi}$
\EndFor
\State \textbf{return} $\pi_\psi$
\end{algorithmic}
\end{algorithm}

\section{Experimental Details}
\label{sec:experimental_details}
\subsection{Simulation Tasks and Demonstrations}
\label{sec:simulation_details}

\begin{figure*}[!t]
    \centering
    \includegraphics[width=\linewidth]{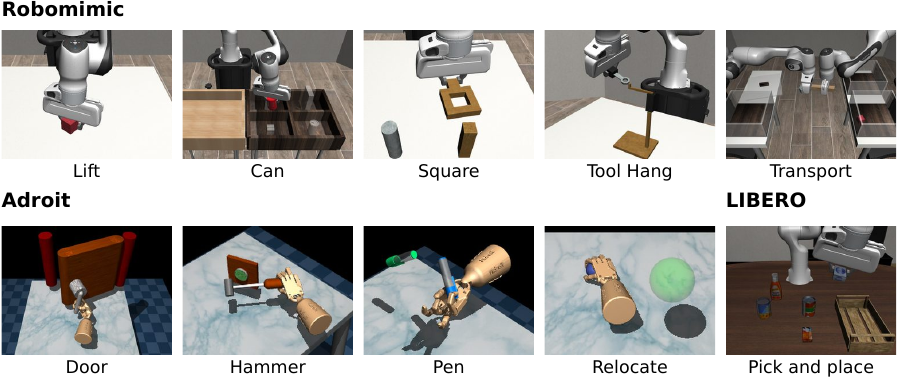}
    \caption{Robomimic (top), Adroit (bottom left), and LIBERO (bottom right) simulation tasks.}
    \label{fig:simulation_tasks}
\end{figure*}

Figure~\ref{fig:simulation_tasks} illustrates the simulation tasks. \textbf{Robomimic} uses state observations comprising robot proprioception and object states, with control at $20$\,Hz. Lift requires lifting a cube, Can requires placing a can in a bin, and Square requires fitting a nut onto a peg. Tool Hang involves assembling a frame and hanging a tool on it, while Transport requires two arms to transfer an object between bins. We train the base policies on $300$ Multi-Human (MH) demonstrations per task for Lift, Can, Square, and Transport, and on $200$ Proficient-Human (PH) demonstrations for Tool Hang~\citep{mandlekar2021matters}.

\noindent\textbf{Adroit} uses state observations and controls a dexterous hand at $100$\,Hz. Its four tasks require opening a door, driving a nail with a hammer, reorienting a pen, and relocating a ball to a target. The DP base policies are trained on the D4RL~\citep{fu2020d4rl} expert-v2 datasets, containing $5{,}000$ trajectories each for Door, Hammer, and Relocate, and $4{,}958$ for Pen.

\noindent\textbf{LIBERO} uses image observations, proprioception, and a language instruction. We use the LIBERO-90 task ``pick up the cream cheese and put it in the tray,'' which runs at $20$\,Hz and provides $50$ human demonstrations~\citep{liu2023libero}. These demonstrations are used for task-specific fine-tuning of the generalist base policies described below.

Across all three benchmarks, ExRL uses a sparse reward of $-1$ at each primitive timestep before success and $0$ upon success, when the episode terminates. Success is determined by the benchmark's task-specific success condition.

\subsection{Policy Architectures and Pretraining}
\label{sec:policy_architectures}

\noindent\textbf{DP and ACT.}
Both policies condition on the current state and predict an $H$-step action chunk. On Robomimic, DP uses a residual MLP denoiser with three hidden layers. The hidden width is $512$ for Lift and Can, $1024$ for Square, $2048$ for Tool Hang, and $4096$ for Transport. On Adroit, DP uses a temporal 1D U-Net with channel widths $(128,256,512)$ and a $128$-dimensional diffusion-step embedding. All DP models use $100$ diffusion steps during training; inference uses $5$ DDIM steps on Lift and Adroit and $20$ on the other Robomimic tasks.

ACT uses a transformer encoder--decoder to predict the action chunk from state inputs. For Lift, Can, and Square, it has a hidden dimension of $512$, $8$ attention heads, $4$ encoder layers, $1$ decoder layer, and a feedforward dimension of $3200$. For Tool Hang and Transport, these values are $768$, $12$, $6$, $1$, and $5120$, respectively. We train ACT without the variational latent encoder and execute its predicted chunks without temporal ensembling.

Table~\ref{tab:base_pretraining_epochs} reports the training epochs of the Robomimic base-policy checkpoints. For Adroit, we use DP checkpoints at epoch $2$ for Door, Hammer, and Pen, and epoch $10$ for Relocate, with $1000$ gradient updates per epoch and a batch size of $1024$.

\begin{table}[!t]
\caption{Training epochs of the Robomimic base policies used to initialize online adaptation. Fractional ACT epochs correspond to checkpoints saved within a data pass.}
\label{tab:base_pretraining_epochs}
\centering
\small
\setlength{\tabcolsep}{4pt}
\renewcommand{\arraystretch}{1.12}
\begin{tabular}{lrrrr}
\toprule
& \multicolumn{2}{c}{DP} & \multicolumn{2}{c}{ACT} \\
\cmidrule(lr){2-3}\cmidrule(lr){4-5}
Task & $H=32$ & $H=50$ & $H=32$ & $H=50$ \\
\midrule
Lift & $20{,}000$ & $10{,}000$ & $0.25$ & $0.12$ \\
Can & $20{,}000$ & $20{,}000$ & $0.92$ & $0.82$ \\
Square & $20{,}000$ & $20{,}000$ & $1190.48$ & $1388.89$ \\
Tool Hang & $4{,}000$ & $28{,}000$ & $100.10$ & $100.07$ \\
Transport & $10{,}000$ & $10{,}000$ & $328.59$ & $250.08$ \\
\bottomrule
\end{tabular}

\end{table}

\noindent\textbf{Generalist base policies.}
On LIBERO, we use the released $\pi_0$ LIBERO checkpoint and task-specific checkpoints of $\pi_{0.5}$, $\pi_0$-FAST, and Cosmos-Policy after $50$, $10{,}000$, and $1500$ fine-tuning updates, respectively. All receive $224\times224$ external-camera and wrist-camera images, proprioception, and the task instruction, and predict chunks of length $H=50$.

\noindent\textbf{ExRL on image observations.}
On LIBERO, ExRL encodes a $64\times64$ external-camera image using four $32$-channel convolutional layers with $3\times3$ kernels, ReLU activations, and strides $(2,1,1,1)$. A fully connected layer projects the flattened features to a $50$-dimensional tanh bottleneck, which is concatenated with proprioception and the flattened predicted chunk. Ten $512\times3$ MLP critic heads share the encoder and output values for each execution length. The SAC actor uses its own encoder and MLP of the same dimensions, with a categorical output. Random crops and color jitter augment images during RL updates.

\subsection{Action-Chunk Length Study}
\label{app:prefix_study}

The left panel of Figure~\ref{fig:intro} examines how action-chunk length affects online adaptation. For convenience, we construct policies with different chunk sizes by exposing only the first $L\in\{2,4,8,16,25,50\}$ actions of one pretrained DP checkpoint with $H=50$ on \texttt{Tool Hang}. Each truncated policy presents the same black-box interface to the baseline RL policies as a base policy with chunk size $L$: actions beyond the prefix are neither executed nor visible to their actors or critics. ResFiT and DICE-RL learn residual corrections only for the exposed $L$ actions, whereas ExRL-SAC selects an execution length $k\leq L$ from the exposed prefix.

\begin{figure}[!t]
\centering
\includegraphics[width=\linewidth]{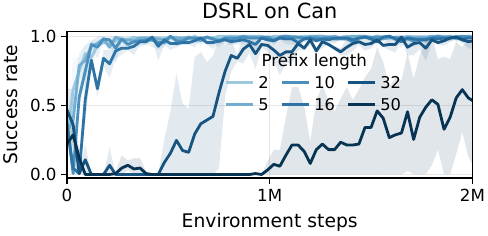}
\caption{\textbf{DSRL on Can.} Three-seed means and $95\%$ CIs. Prefixes $2$--$32$ share an $H=32$ checkpoint; prefix $50$ uses an $H=50$ checkpoint.}
\label{fig:dsrl_can_prefix}
\end{figure}
For DSRL, we also restrict the learnable latent space to the prefix. Its actor outputs an $L\times d_a$ noise tensor, where $d_a=7$ is the primitive action dimension. Inside the base-policy wrapper, the remaining $(H-L)\times d_a$ noise entries are independently sampled from a clipped Gaussian. The original diffusion policy then decodes the complete $H$-step chunk, but the wrapper returns only the first $L$ actions. The action critic receives only this action prefix, and the noise critic receives only the actor's $L$-step noise; neither sees the hidden suffix. The pretrained diffusion model remains unchanged, while DSRL's learned control dimension scales with $L$. All variants use $20$ DDIM steps and a task horizon of $800$ primitive steps. We report three-seed means and $95\%$ confidence intervals.

\begin{figure*}[!t]
      \centering
      \includegraphics[width = 1.0\linewidth]{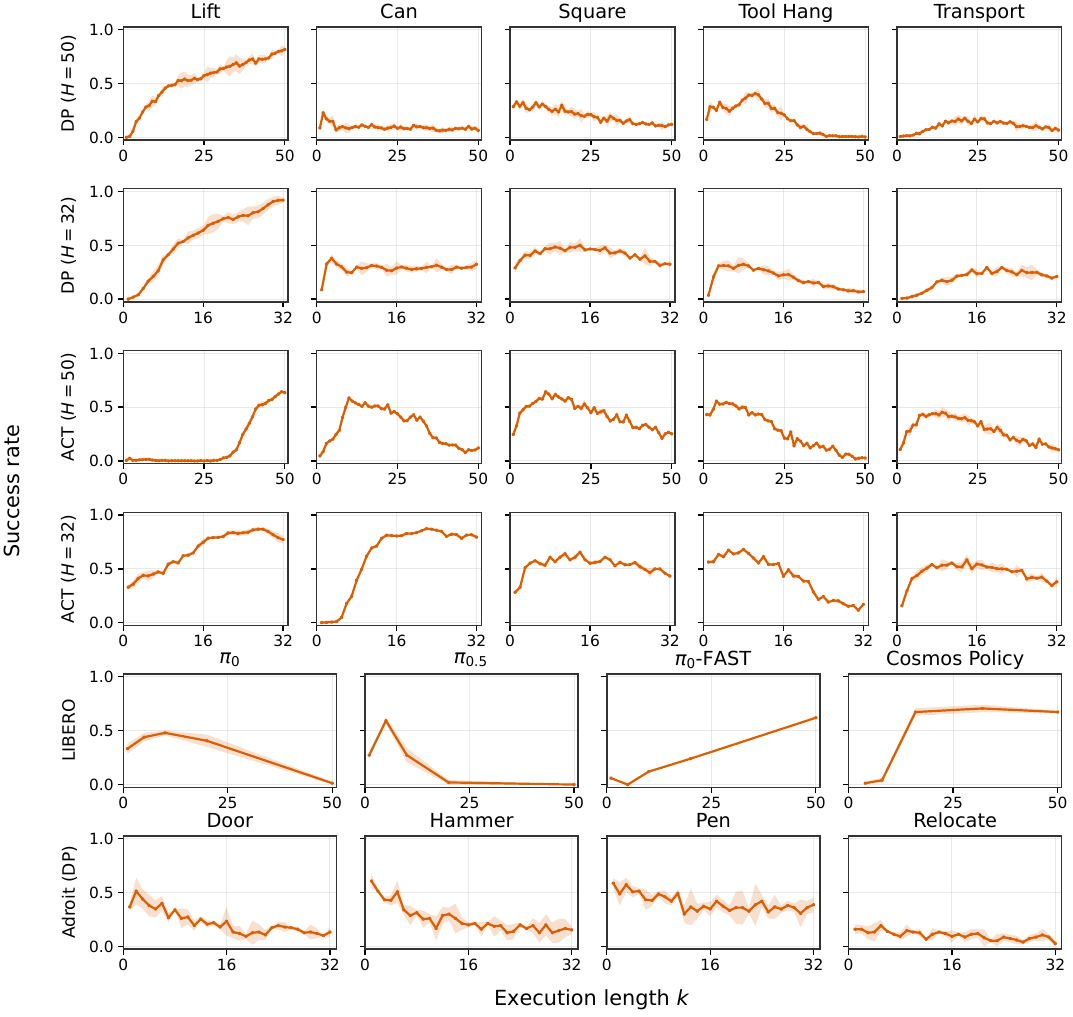}
      \caption{\textbf{Fixed-$k$ sweep across base policies and tasks.}
We evaluate each base policy under different fixed execution lengths on \texttt{Robomimic}, \texttt{LIBERO}, and \texttt{Adroit}. Performance varies substantially with $k$, and the optimal fixed execution length differs across tasks and policies, highlighting the importance of adaptive execution-length control. Shaded bands indicate 95\% confidence intervals over three seeds.}
      \label{fig:all_fixed_K}
\end{figure*}

The frozen DP on \texttt{Tool Hang} is itself less successful when executing long prefixes (Figure~\ref{fig:all_fixed_K}). To separate the effect of chunk length on RL adaptation from differences in base-policy performance, we additionally compare DSRL on \texttt{Can}, where DP has similar success rates across prefixes. As shown in Figure~\ref{fig:dsrl_can_prefix}, DSRL quickly reaches near-perfect success with prefixes of $2$--$16$ actions, but learns substantially more slowly with $32$ and $50$ actions. The sample-efficiency gap thus persists even when the base policy performs similarly at different prefix lengths.

\subsection{Shared Hyperparameters}
\label{sec:exrl_q_hyperparameters}
\label{sec:exrl_hyperparameters}

Most settings of ExRL are shared across tasks and base policies. Table~\ref{tab:exrl_shared_hparams} gives the defaults for ExRL-Q and ExRL-SAC, and Table~\ref{tab:exrl_task_hparams} lists the main task-specific changes. The same RL settings apply when Online BC is enabled.

\begin{table}[!t]
\caption{Shared ExRL hyperparameters.}
\label{tab:exrl_shared_hparams}
\centering
\small
\setlength{\tabcolsep}{4pt}
\renewcommand{\arraystretch}{1.08}
\begin{tabular}{lcc}
\toprule
Parameter & ExRL-Q & ExRL-SAC \\
\midrule
Optimizer & Adam & Adam \\
Critic learning rate & $3\times10^{-4}$ & $3\times10^{-4}$ \\
Critic hidden layers & $512\times3$ & $512\times3$ \\
Activation & ReLU & ReLU \\
Critic ensemble size & $10$ & $10$ \\
Target critic subsample size & $2$ & $2$ \\
Target update rate $\tau$ & $0.005$ & $0.005$ \\
Batch size & $256$ & $256$ \\
Replay capacity (transitions) & $10^6$ & $10^6$ \\
Primitive discount $\gamma$ & $0.997$ & $0.997$ \\
CQL coefficient & $0.3$ & $0.01$ \\
CQL temperature & $1$ & $1$ \\
TD target clipping & $[-1000,0]$ & $[-1000,0]$ \\
UTD & $1/4$ & $1$ \\
Warmup size & $512$ & $512$ \\
Updates on warmup replay & $5{,}000$ (critic) & $5{,}000$ (actor--critic) \\
\midrule
Actor hidden layers & --- & Same as critic \\
Actor learning rate & --- & $5\times10^{-5}$ \\
Initial entropy temperature & --- & $0.3$ \\
Temperature learning rate & --- & $10^{-5}$ \\
Target entropy & --- & $0.98\log H$ \\
\bottomrule
\end{tabular}

\end{table}

\begin{table}[!t]
\caption{Main task-specific hyperparameter changes.}
\label{tab:exrl_task_hparams}
\centering
\small
\setlength{\tabcolsep}{4pt}
\renewcommand{\arraystretch}{1.14}
\begin{tabular}{p{0.36\linewidth}p{0.57\linewidth}}
\toprule
Setting & Changes to the shared defaults \\
\midrule
Transport, all base policies & Critic and SAC actor hidden layers: $2048\times3$. \\
Transport, DP; ACT with ExRL-SAC & $\gamma=0.999$. \\
Tool Hang, all base policies & $\gamma=0.999$. \\
LIBERO, all base policies & Replay capacity: $10^5$; ExRL-Q UTD: $1$. \\
\bottomrule
\end{tabular}

\end{table}

Default episode horizons are $300$ for Lift and Can, $400$ for Square, $800$ for Tool Hang and Transport, $200$ for Adroit, and $250$ for LIBERO. Base-policy architectures and the LIBERO image encoder are described in Appendix~\ref{sec:policy_architectures}.

\subsection{Online BC and Demonstration Augmentation}
\label{sec:bc_hyperparameters}

Table~\ref{tab:online_bc_hparams} summarizes the default base-policy update settings. Successful trajectories are sampled uniformly over episodes, and BC starts once enough successful experience is available. For DP and ACT, we add a teacher constraint to the online BC loss using a frozen copy of the initial base policy. DP minimizes the mean squared difference between student and teacher noise predictions at identical observations, noisy chunks, and diffusion timesteps; ACT matches their inference-time action chunks at identical observations. The teacher stays fixed throughout RL, limiting drift from pretrained behavior as the student learns from successful online trajectories.

\begin{table}[!t]
\caption{Online BC hyperparameters. DP settings apply to both Robomimic and Adroit.}
\label{tab:online_bc_hparams}
\centering
\small
\setlength{\tabcolsep}{4pt}
\renewcommand{\arraystretch}{1.12}
\begin{tabular}{lcc}
\toprule
Parameter & DP & ACT \\
\midrule
BC learning rate & $10^{-5}$ & $10^{-5}$ \\
BC batch size & $256$ & $128$ \\
BC updates / RL update & $1/2$ & $1/32$ \\
Success-buffer capacity (windows) & $25{,}000$ & $25{,}000$ \\
Teacher weight & $1$ & $1$ \\
\bottomrule
\end{tabular}

\end{table}

DP Tool Hang delays BC until $200$k environment steps. ACT uses a gradient-norm limit of $10$.

\noindent\textbf{Demonstration augmentation.}
The default demonstration fraction in each BC batch is $0.5$, and the frozen-teacher constraint is disabled. The number of BC updates per RL update decreases linearly from $5$ to $1$ over the first $600$k environment steps after BC begins, and remains at $1$ thereafter.

\noindent\textbf{Demonstrations in RL replay.}
For the \texttt{Can} comparison in Figure~\ref{fig:offline_human_demos}, the critic batch contains $20\%$ relabeled demonstration transitions and $80\%$ online transitions. The demonstration buffer holds up to $2.5$M transitions, separately from the online replay.

\subsection{Baseline Implementations and Settings}
\label{sec:baseline_settings}

We adapt the released baselines to our policy backbones and chunk sizes. Table~\ref{tab:baseline_tuning} summarizes hyperparameter selection; below, we describe changes to their policy interfaces and learning procedures.

\begin{table*}[!t]
\caption{Hyperparameter selection for the RL baselines.}
\label{tab:baseline_tuning}
\centering
\small
\setlength{\tabcolsep}{4pt}
\renewcommand{\arraystretch}{1.12}
\begin{tabular}{@{}p{0.18\linewidth}p{0.78\linewidth}@{}}
\toprule
Baseline & Hyperparameter selection \\
\midrule
DSRL & Task-specific searches over the DSRL-NA/SAC variant, latent-noise range, learning rate, update frequency, discount, and network width. \\
Policy Decorator & Task-specific searches over residual scale, entropy coefficient, learning rates, progressive exploration, and actor/critic update frequencies. \\
PA-RL & Searches over critic capacity and update frequency, action-optimization steps and step size, and the learning rate, start time, and frequency of policy distillation. \\
ResFiT & Search on DP \texttt{Square}, then reuse the selected optimizer, residual, exploration, and update settings across tasks and backbones. \\
OGPO & Adapt the released learning objectives to DP and adjust update frequency for our collection setup and longer chunks. \\
DICE-RL & Use the paper's Can recipe for the main learning hyperparameters, with adaptations to our policies, replay, and demonstration datasets. \\
\bottomrule
\end{tabular}

\end{table*}

\noindent\textbf{DSRL.}
The released Robomimic configurations of \citet{DSRL} typically use chunks of $4$ or $8$ actions. We connect the official learner to our DP checkpoints and extend its latent action space to $H$-step chunks. The diffusion sampler and normalization follow each checkpoint. We consider both DSRL-NA and DSRL-SAC during tuning. For example, the selected DP \texttt{Transport} and \texttt{Tool Hang} configurations use DSRL-SAC, a latent-noise bound of $0.75$, and a learning rate of $5\times10^{-5}$, with network widths of $2048$ and $512$, respectively. The restricted latent interface used in the prefix study is described in Appendix~\ref{app:prefix_study}.

\noindent\textbf{Policy Decorator.}
We port the residual SAC learner of \citet{Policy-decorator} to our training framework while retaining bounded residual actions and progressive exploration. The residual has the same shape as the executed action chunk, and the actor conditions on both the observation and the base chunk, using the observation-plus-base-action option of the official code instead of its observation-only default. For DP with $H=50$, the selected residual scales are $0.1$, $0.15$, $0.05$, $0.02$, and $0.02$ on \texttt{Lift}, \texttt{Can}, \texttt{Square}, \texttt{Tool Hang}, and \texttt{Transport}, respectively. The two long-horizon tasks also use smaller entropy coefficients.

\noindent\textbf{PA-RL.}
We implement \citet{PA-RL} in our PyTorch framework, replacing its original base-policy interface with our DP and ACT models. Action optimization operates on complete chunks, and distillation uses the corresponding native training objective: denoising for DP and action reconstruction for ACT. We retain the official distributional critic ensemble and Cal-QL regularization, with $10$ critics, $2$-critic target subsampling, and regularization coefficient $0.005$. Our main changes concern the interaction between critic learning, action optimization, and policy distillation. We first warm up the critic, then collect experience with optimized actions and periodically distill them into the base policy. For DP, we sample $32$ candidates and retain $10$ for optimization. ACT retains its deterministic inference, so repeated base candidates coincide; local optimization and the inclusion of replay actions provide the action-improvement step. Local search uses $10$--$20$ gradient steps, with its step size and the distillation schedule tuned by task. On DP \texttt{Lift}, distillation uses globally selected candidates without local refinement, whereas rollout actions still receive local optimization. \texttt{Lift}, \texttt{Can}, and \texttt{Square} use online replay only. The $H=50$ \texttt{Transport} and \texttt{Tool Hang} runs additionally mix equal proportions of offline demonstrations and online experience, following the official mixed-data recipe. Reported performance evaluates the optimized-action policy.

\noindent\textbf{ResFiT.}
We extend the primitive-action residual formulation of \citet{ankile2025residual} to complete action chunks: the residual actor receives the state and base chunk, and the critic evaluates the composed chunk. Replay stores one transition per executed chunk, with accumulated discounted rewards and a bootstrap discount determined by its actual duration. We retain the TD3-style updates, $10$-critic ensemble, bounded residuals, and equal mixing of demonstration and online data. After searching on DP \texttt{Square}, we reuse actor/critic learning rates of $5\times10^{-6}$/$10^{-4}$, residual scale $0.1$, exploration/target noise standard deviations of $0.025$/$0.01$, and target update rate $0.01$. Each collected chunk transition receives one critic update and one actor update every four critic updates. We use one-chunk TD targets instead of the released primitive-step multi-step targets. Network width and discount remain task-specific. The same learner is used with ACT and Adroit; for LIBERO, we add image encoders and connect it to each policy's native observation and action transforms.

\noindent\textbf{OGPO.}
The released implementation of \citet{ogpo} uses flow-matching policies in JAX. To fine-tune our pretrained DP models directly, we implement its learning objectives in PyTorch and construct stochastic denoising trajectories around the DP's DDIM transitions. We retain OGPO's clipped likelihood-ratio objective and conservative group advantages, evaluating old and current likelihoods on the same denoising trajectory. We replace flow-matching regularization with the DP denoising loss on successful online experience. The standard runs use $32$ samples per advantage group and best-of-$8$ action selection during collection; evaluation samples a single trajectory from the target policy. We use lower update frequencies than the released configuration, with task-specific schedules for parallel collection and long chunks.

\noindent\textbf{DICE-RL.}
We adapt \citet{DICE-RL} to our frozen DP and ACT policies while retaining its noise-conditioned residual, critic-guided candidate selection, and selective behavior regularization. For DP, the same noise conditions the base policy and residual actor. For ACT, the base chunk uses its standard deterministic inference, while noise conditions only the residual actor. We use the paper's Can settings: $10$ critics, $16$ rollout/actor candidates, $4$ target candidates, behavior-regularization weight $50$, and an offline replay fraction that decreases from $0.7$ to $0.1$ over $160$k environment steps. Relative to the released Can configuration, we enable the paper's candidate-level Q filter and use constant learning rates of $10^{-4}$. Demonstrations come from the datasets used for our base policies. Our replay stores non-overlapping chunk transitions, replacing the release's overlapping action windows; three-step TD targets therefore span three chunk decisions, with rewards summed within each chunk and discounting applied between decisions. The RL batch size is $256$. Robomimic uses $2.5$ critic updates per collected chunk transition, and Adroit uses one; the actor and target networks update half as frequently.

\noindent\textbf{AAC.}
We adapt the released inference-time AAC selector~\citep{AAC} to each policy's native chunk size. Robomimic and LIBERO use $20$ candidate chunks and the original entropy-increase rule with its motion constraint. On bimanual \texttt{Transport}, we sum the two arms' entropies and average their motion measures. Adroit uses an entropy-only adaptation over joint actions with $64$ candidates, since the original Cartesian motion constraint does not apply. For $\pi_0$-FAST, auxiliary candidates use token sampling at temperature $1$ to estimate entropy, while the executed candidate retains greedy decoding. In every case, AAC selects a prefix without averaging or re-ranking the candidate actions.

\noindent\textbf{Other non-RL baselines.}
Random-$k$ samples uniformly from $\{1,\ldots,H\}$ at each query, and optimal fixed-$k$ reports the best fixed execution length from a sweep. Online BC executes full chunks and learns only from successful online trajectories, using the imitation objective and teacher constraint described in Appendix~\ref{sec:bc_hyperparameters}, without RL updates or a learned execution-length policy.

\subsection{Details of Real-World Experiment}
\label{sec:real_world_details}
All real-world experiments use an AgileX Piper arm controlled at $20$\,Hz. Each task uses a separate $\pi_{0.5}$ base policy fine-tuned on human demonstrations. The policy receives third-person and wrist RGB views together with proprioceptive state, and predicts $H=50$ actions. ExRL-Q keeps this policy frozen and selects the execution length using the mean of ten critics. Table~\ref{tab:real_world_training_details} summarizes the training settings.

\begin{table*}[!t]
\caption{Real-world ExRL-Q settings. Rollout totals include warmup. Only the execution-length critic is updated during RL.}
\label{tab:real_world_training_details}
\centering
\footnotesize
\setlength{\tabcolsep}{4pt}
\renewcommand{\arraystretch}{1.08}
\begin{tabular}{lcccc}
\toprule
Setting & Stack cube & Clothing folding & Power insertion & Parts organization \\
\midrule
BC demonstrations & $13$ & $50$ & $128$ & $21$ \\
BC frames & $2{,}995$ & $38{,}412$ & $22{,}865$ & $19{,}288$ \\
Warmup rollouts & $25$ & $14$ & $30$ & $29$ \\
Offline critic updates & $10{,}000$ & $10{,}000$ & $5{,}000$ & $5{,}000$ \\
Online updates & UTD $16$ & UTD $16$ & $400$/rollout & $400$/rollout \\
Critic hidden dimensions & $(512,512,512)$ & $(512,512,512)$ & $(1024,1024,1024)$ & $(1024,1024,1024)$ \\
Batch size & $256$ & $256$ & $64$ & $64$ \\
Discount $\gamma$ & $0.997$ & $0.999$ & $0.997$ & $0.997$ \\
Training rollout limit & $16$\,s & $60$\,s & $450$ steps & $1021$ steps \\
Plotted rollouts & $43$ & $93$ & $141$ & $250$ \\
\bottomrule
\end{tabular}

\end{table*}

\begin{figure}[!t]
\centering
\includegraphics[width=0.7\linewidth]{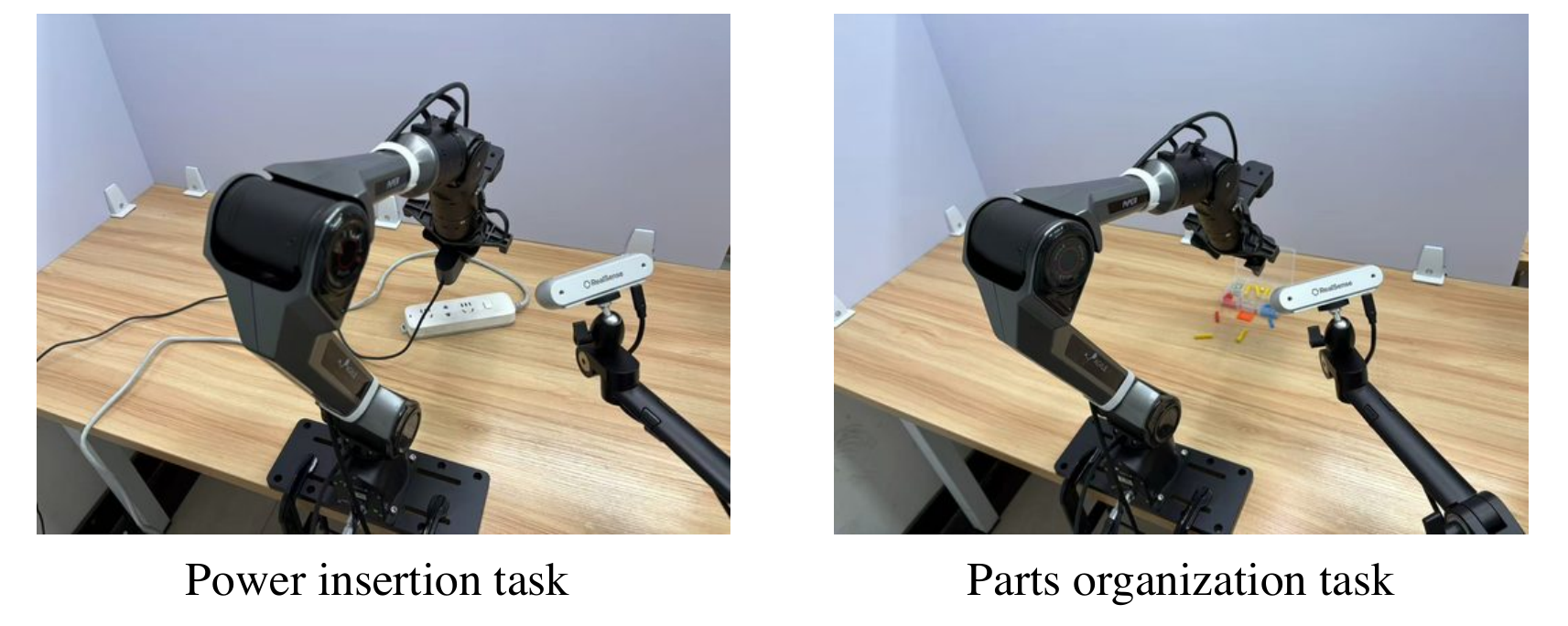}
\caption{Example setups for power insertion and parts organization.}
\label{fig:setup}
\end{figure}

\noindent\textbf{Tasks and interaction.}
Our initial experiments study power insertion and parts organization (Figure~\ref{fig:setup}). Insertion requires placing a charger plug into a power strip; organization requires placing four parts into separate compartments and closing the lid, giving five progress stages. We later extend the evaluation to stacking one cube on another and folding a T-shirt through three successive stages. During RL, the operator marks task or stage completion without providing action corrections or teleoperation.

\noindent\textbf{Accounting for the cost of replanning.}
The initial insertion and organization experiments revealed a practical distinction between learning in fewer robot actions and completing a rollout quickly. Early in RL, the adapter often favored short execution lengths, requiring frequent queries to the large base policy and slowing each rollout. The original rewards did not directly penalize these inference pauses: insertion received $-1$ per primitive timestep before success and $0$ at success, while organization received $+1$ for each completed stage and zero otherwise.

This observation motivated the elapsed-time objective used in our later stacking and folding experiments. Both the reward and discount now account for physical time, including inference within a rollout, so the adapter must balance the benefit of replanning against its time cost. For a decision lasting $\Delta$ seconds, the bootstrap discount is $\gamma^{\Delta/0.05}$, and the reward accumulates a discounted cost over the same interval. Stacking incurs unit cost per $50$-ms reference interval until success; folding weights this cost by the number of unfinished stages. The latter rewards both stage completion and timely progress through the remaining fold. Successful rollouts terminate, while ordinary time limits retain bootstrapping.

\noindent\textbf{Policy preparation and training.}
The stacking and folding base policies are fine-tuned for $900$ and $10{,}000$ gradient steps, respectively, with folding updating the action expert. Stacking uses a D455 third-person camera and a D405 wrist camera; folding uses two D405 cameras. Folding views are cropped to $360\times360$ before base-policy preprocessing and resized to $128\times128$ for the critic. All tasks use critic learning rate $3\times10^{-4}$, CQL weight $0.3$, and two sampled target critics. We initialize the critic from warmup data before online adaptation; stacking and folding collect at least $256$ and $512$ transitions, respectively. Demonstration counts, update schedules, and rollout limits are given in Table~\ref{tab:real_world_training_details}.

\noindent\textbf{Learning curves and evaluation.}
Figure~\ref{fig:real_world_adaptation} shows an EMA with coefficient $0.2$, initialized at zero and updated after each rollout, including warmup. The outcome is binary success for stacking, folding, and insertion, and the fraction of completed stages for organization. For stacking and folding, the horizontal axis sums physical rollout time, including inference pauses but excluding resets, learner updates, and waiting between rollouts. The curves cover $10.07$ and $84.06$ minutes, respectively; insertion and organization use episode counts. Final evaluations are reported separately in Table~\ref{tab:real_robot_eval}.

\section{On-Policy versus Off-Policy Adaptation}
\label{app:ppo_comparison}

To examine how the choice of RL algorithm affects sample efficiency, we compare ExRL-Q with a PPO execution-length adapter, motivated by the PPO-based approach of \citet{zhao2026dynamic}. Both adapters use the same frozen DP checkpoint on \texttt{Can}, with chunk size $H=50$, the same state and action-chunk inputs, and the same execution-length choices.

\begin{figure}[!t]
    \centering
    \includegraphics[width=0.94\linewidth]{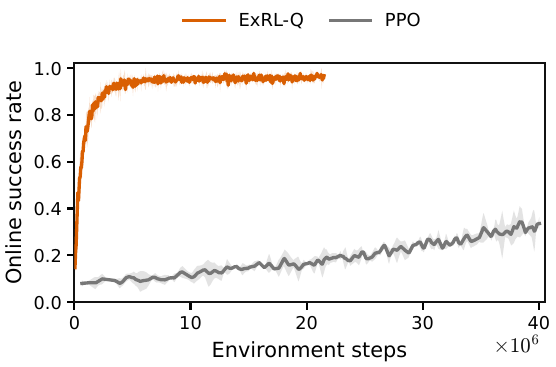}
    \caption{\textbf{PPO versus ExRL-Q on Can.} Mean online success over $200$ episodes, with $95\%$ CIs.}
    \label{fig:ppo_comparison}
\end{figure}
As shown in Figure~\ref{fig:ppo_comparison}, ExRL-Q reaches over $90\%$ success within approximately $3$M environment steps, while PPO remains below $40\%$ after approximately $40$M steps. Curves average success over the most recent $200$ training episodes; the horizontal axis counts primitive environment steps. This substantial gap supports the use of off-policy learning for sample-efficient execution-length adaptation.

\begin{figure*}[!t]
      \centering
      \includegraphics[width = 1.0\linewidth]{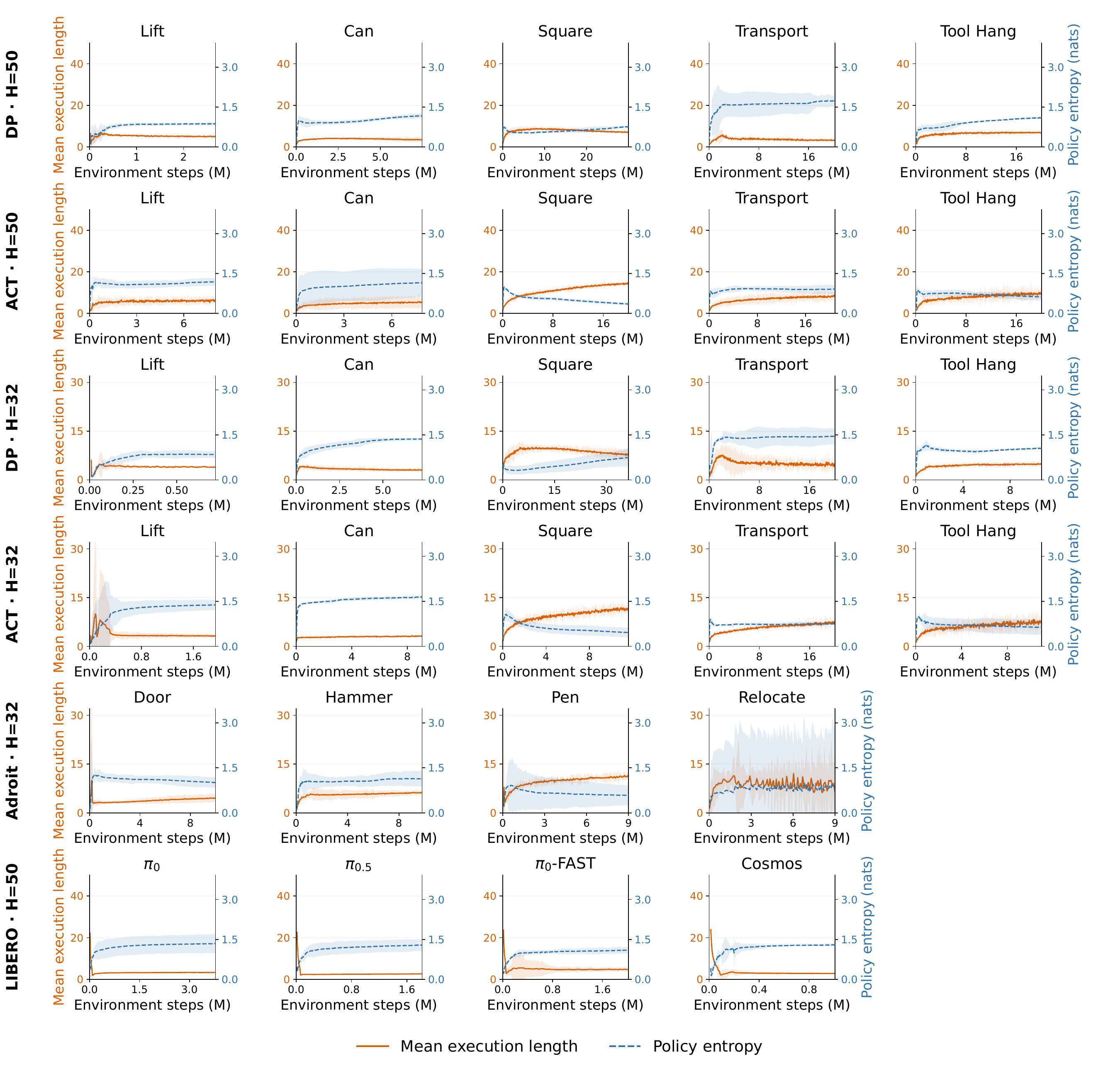}
      \caption{\textbf{Evolution of execution lengths and policy entropy learned by ExRL-SAC.}
Orange solid curves show mean execution length; blue dashed curves show the categorical policy entropy on training minibatches. Shaded bands indicate $95\%$ confidence intervals over three training seeds.}
      \label{fig:avg_k}
\end{figure*}

\section{Evolution of Learned Execution Lengths}

Figure~\ref{fig:avg_k} examines how execution-length strategies develop during online adaptation. Alongside the mean execution length, we plot the categorical policy entropy averaged over training minibatches, which measures how concentrated the adapter's decisions are for a given state and predicted chunk. Across tasks and base policies, the learned mean remains well below the full chunk size, but its evolution differs substantially. All four \texttt{LIBERO} policies rapidly settle at approximately $3$--$5$ actions, whereas ACT on \texttt{Square} gradually favors longer execution, reaching about $14$ actions with $H=50$ and $11$ with $H=32$. Adroit \texttt{Pen} exhibits a similar increase.

Policy entropy further distinguishes these strategies. On ACT \texttt{Square} and Adroit \texttt{Pen}, increasing execution length accompanies decreasing entropy: the adapter becomes more decisive while allowing longer motions. In contrast, the \texttt{LIBERO} policies retain nonzero entropy after their mean execution lengths stabilize, so frequent replanning does not require deterministic length selection. The same adapter can therefore learn sustained frequent replanning or an increasing preference for longer motions, depending on the task and base policy.

\section{Additional Execution-Length Analysis}
\label{app:execution_length_analysis}

The behaviors in Section~\ref{sec:execution_patterns} suggest that ExRL benefits from choosing when to seek a new prediction and when to continue an existing one. Here we examine this interpretation beyond the main-text examples. We first establish that the execution patterns recur across tasks, then separate the effect of choosing suitable lengths from choosing them at suitable moments. Finally, we examine why frequent replanning can sometimes interrupt progress. The \texttt{Robomimic} analyses use frozen base policies with $H=32$; the real-world examples use $H=50$.

\subsection{Execution Patterns across Tasks}
\label{app:phase_statistics}

The tendency to shorten execution near interaction is consistent across tasks and random seeds. Figure~\ref{fig:visulization}(a) compares free motion with the approach to an object or insertion target, using the same trajectories for both phases. ExRL selects shorter lengths near interaction in all nine simulation task--seed pairs as well as on the real robot. This pattern persists when the distance threshold is changed from $8$\,cm to $6$ or $10$\,cm, suggesting that it reflects a recurring behavioral preference rather than a particular choice of phase boundary.

\begin{figure*}[!t]
\centering
\includegraphics[width=\linewidth]{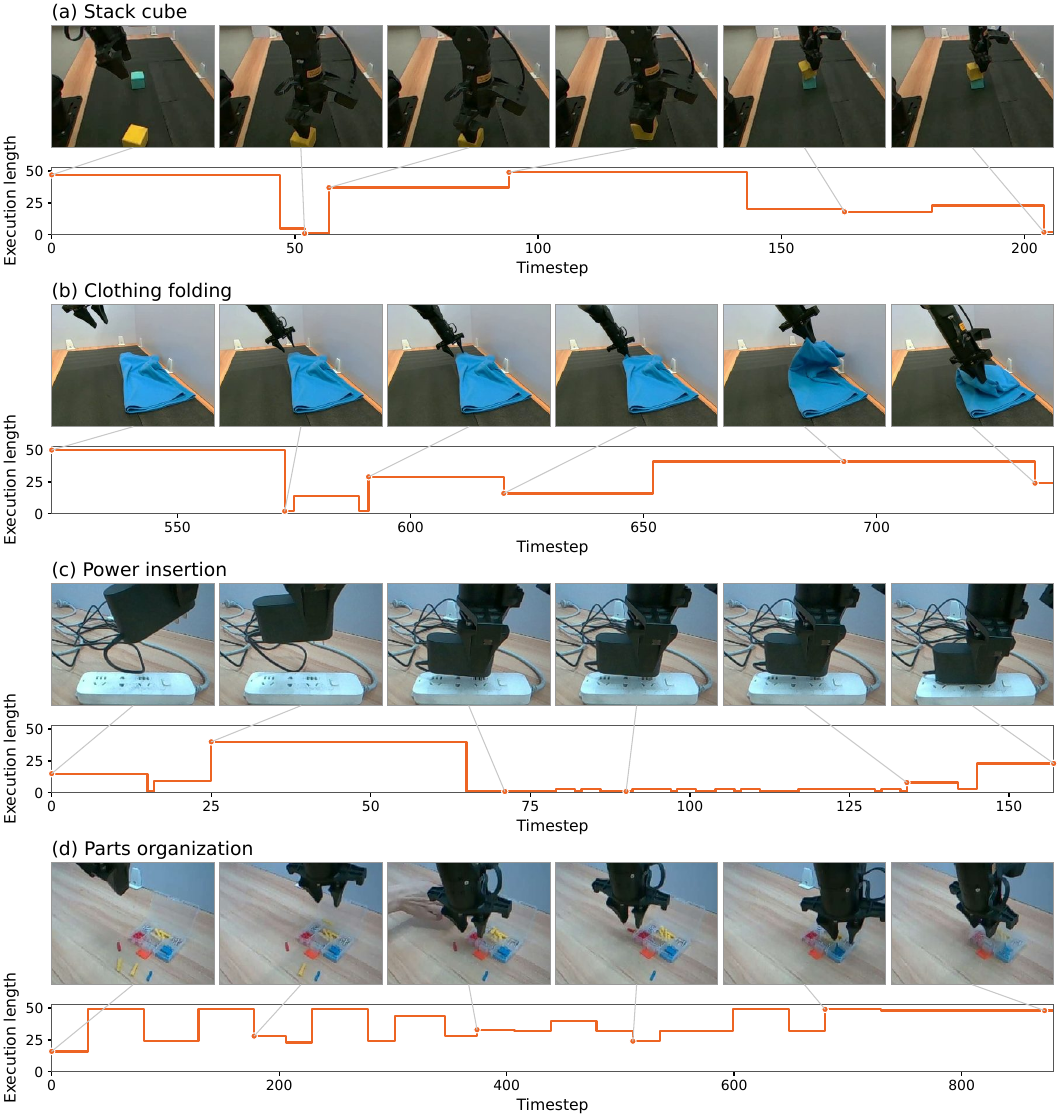}
\caption{\textbf{ExRL varies execution length within a rollout to balance local adjustments and sustained motion.} Representative frames and execution-length decisions across four real-world tasks. For example, stacking and folding use short executions near grasping, followed by longer motions to lift and transport the object or fabric; insertion concentrates short executions near contact, whereas parts organization sustains longer executions across much of the pick-and-place sequence.}
\label{fig:real_execution_patterns}
\end{figure*}

The real-world examples in Figure~\ref{fig:real_execution_patterns} show how this preference fits into a complete behavior. Stacking and folding use short adjustments around grasping before continuing with longer motions, while insertion concentrates replanning near contact. Parts organization, by contrast, sustains relatively long executions across much of its pick-and-place sequence. Repeated object interactions therefore do not necessarily require uniformly short execution: the appropriate length also depends on how effectively the predicted chunk carries the task forward.

For the phase comparison, we collect $50$ rollouts per simulation task and seed, and $25$ real insertion rollouts. We retain trajectories that reach interaction and contain both phases: $86$ for \texttt{Lift}, $145$ for \texttt{Can}, $121$ for \texttt{Square}, and $21$ for insertion. Queries before the first grasp or insertion are split by their starting distance to the object or target. We average selected lengths within each phase and then across trajectories; confidence intervals bootstrap trajectories.

\subsection{Matching Lengths to States and Chunks}
\label{app:conditional_assignment}

A favorable mixture of short and long executions might explain some of ExRL's gains even without adapting to the current situation. We isolate this possibility on \texttt{Square} by keeping the selected lengths unchanged but shuffling their assignment across environments at each synchronized query. The control receives exactly the same multiset of lengths as ExRL-Q, while losing the correspondence between each length and the state--chunk pair for which it was chosen.

This reassignment reduces success by $38$--$58$ percentage points across DP and ACT (Table~\ref{tab:conditional_assignment}). The benefit therefore lies not only in which lengths are used, but in when they are used. This result supports the interpretation of the main-text examples: short corrections, rejection of unsuitable chunks, and longer execution are useful when matched to the current state and prediction.

\begin{table}[tbp]
\centering
\caption{\textbf{Conditional assignment matters beyond the length distribution.} \texttt{Square} success rates (counts), and gains over the permuted control in percentage points with 95\% cluster-bootstrap intervals.}
\label{tab:conditional_assignment}
\small
\setlength{\tabcolsep}{4pt}
\renewcommand{\arraystretch}{1.12}
\begin{tabular}{lccc}
\toprule
Base policy & ExRL-Q & \makecell{Permuted\\lengths} & Gain (95\% CI) \\
\midrule
DP & 95.8\% (23/24) & 37.5\% (9/24) & $58.3\;[37.5,79.2]$ \\
ACT & 70.8\% (17/24) & 33.3\% (8/24) & $37.5\;[16.7,58.3]$ \\
\bottomrule
\end{tabular}
\end{table}

We use three independently trained ExRL-Q checkpoints per base policy. Paired rollouts share the initial state, initial chunk, and policy-inference randomness; confidence intervals resample checkpoint--environment-slot clusters.

\begin{figure*}[!t]
\centering
\includegraphics[width=\linewidth]{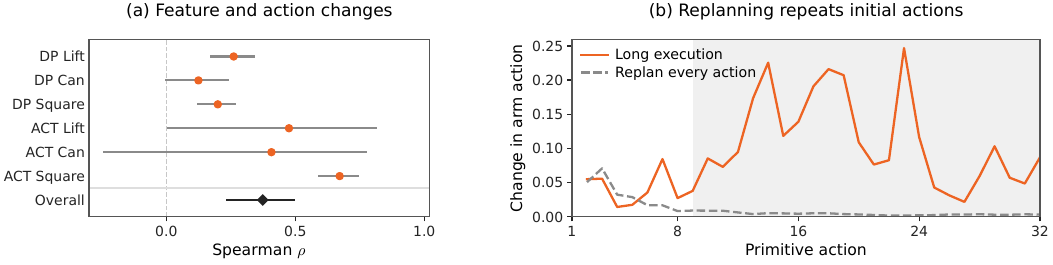}
\caption{\textbf{Frequent replanning can repeat the start of a useful action sequence.}
(a) Feature changes correlate with changes in the newly predicted first action across DP and ACT ($95\%$ CIs). (b) From the same state and initial chunk, replanning produces nearly unchanged commands, while longer execution preserves the changing actions within the chunk.}
\label{fig:feature_aliasing}
\end{figure*}

\subsection{Feature Aliasing and Repeated Replanning}
\label{app:feature_aliasing}

Short executions help when a new prediction provides a useful correction or alternative motion. But a new prediction can also restart the same behavior. Each query returns to the first action of a newly predicted chunk; if successive states produce similar policy inputs or internal features, these first actions may remain nearly unchanged. We refer to this local ambiguity as \emph{feature aliasing}. In such situations, continuing through an existing chunk can preserve a progression that repeated replanning fails to recover.

This connection appears across both policy architectures. Over $32$-action replanning windows from $72$ states, smaller feature changes tend to accompany smaller changes in the newly predicted first arm action. The association is positive for DP and ACT on each of \texttt{Lift}, \texttt{Can}, and \texttt{Square} (Figure~\ref{fig:feature_aliasing}(a)), with overall Spearman correlation $\rho=0.374$ ($95\%$ CI $[0.237,0.494]$). A shorter, $16$-action window gives a similar correlation of $0.383$.

Physical motion alone need not break this repetition. Figure~\ref{fig:feature_aliasing}(b) compares two continuations from the same ACT \texttt{Square} state and initial chunk. In the shaded interval, replanning after every action moves the end effector by $19.4$\,mm, yet successive arm commands change by only $3.6\%$ as much as those along the original chunk. The robot moves, but the repeated predictions do not recover the later actions already available in that chunk. This example helps explain why allowing a useful prediction to unfold can be beneficial even when fresh observations are available.

Related analysis by \citet{lazzati2026does} shows that conditioning on past observations can represent non-Markovian demonstrator behavior and mitigate compounding error, with these benefits also obtainable through delayed policies. Our observations provide a complementary perspective: while current features remain locally ambiguous, continuing a chunk may preserve behavioral progression that newly predicted first actions fail to provide.

Preserving this progression still requires choosing how long to continue. Table~\ref{tab:aliasing_continuation} compares three continuations from the same selected ACT \texttt{Square} states: persistent one-action replanning, one full chunk followed by ExRL-Q, and ExRL-Q throughout. The adaptive strategy completes a case in which both alternatives fail. These diagnostic examples support a balance between obtaining new predictions when they help and retaining an existing prediction while it remains useful.

\begin{table}[tbp]
\centering
\caption{\textbf{Completion from states with repeated first actions.} Success counts from the same three selected ACT \texttt{Square} states; these are diagnostic cases, not benchmark success rates.}
\label{tab:aliasing_continuation}
\small
\setlength{\tabcolsep}{4pt}
\renewcommand{\arraystretch}{1.12}
\begin{tabular}{lccc}
\toprule
Base policy / task & \makecell{Always\\$k=1$} & \makecell{$k=32$, then\\ExRL-Q} & ExRL-Q \\
\midrule
ACT / \texttt{Square} & 1/3 & 1/3 & 2/3 \\
\bottomrule
\end{tabular}
\end{table}

\noindent\textbf{Measurement details.}
The ACT feature is the first decoder token; for DP, which has no learned observation encoder, we use the normalized conditioning input. We measure feature changes by normalized $\ell_2$ distance and arm-action changes by $\ell_2$ distance excluding the gripper. Window-level correlations are combined in Fisher-$z$ space, with equal weight across episodes and model--task pairs at the corresponding aggregation levels; confidence intervals resample episodes. The paired example uses the same initial state and chunk for both branches, and the three continuation states were selected before observing their outcomes.

\section{Execution-Length Control as an Induced SMDP}
\label{app:smdp}

In this section, we formalize the connection between ExRL and semi-Markov
decision processes (SMDPs)~\citep{SMDP}. Recall that the underlying environment is
a primitive-time MDP
$\mathcal{M}=(\mathcal{S},\mathcal{A},P,r,\rho_0,\gamma)$, where each primitive
action advances the environment by one timestep. ExRL does not change this
environment. Instead, by freezing an action-chunked policy
$\pi_{\mathrm{base}}$ and learning only how many actions from each predicted
chunk to execute, ExRL induces a higher-level decision process over
execution-length decisions.

Let $\mathbf{a}=a_{t:t+H}$ denote an action chunk sampled from the frozen base
policy at state $s$, i.e.,
$\mathbf{a}\sim\pi_{\mathrm{base}}(\,\cdot\,\mid s)$. The high-level decision
state is the pair
\[
    x=(s,\mathbf{a})\in\mathcal{X}:=\mathcal{S}\times\mathcal{A}^H,
\]
and the high-level action is the execution length
$k\in\mathcal{K}=\{1,\ldots,H\}$. Thus, ExRL induces an SMDP
\[
    \mathcal{M}_{\mathrm{ex}}
    =
    (\mathcal{X},\mathcal{K},P_{\mathrm{ex}},r_{\mathrm{ex}},
    F_{\mathrm{ex}},\rho_{\mathrm{ex}},\gamma),
\]
where $\mathcal{X}$ is the decision-state space, $\mathcal{K}$ is the
execution-length action space,
$P_{\mathrm{ex}}:\mathcal{X}\times\mathcal{K}\rightarrow\Delta(\mathcal{X})$
is the transition kernel over execution-level decision states,
$r_{\mathrm{ex}}:\mathcal{X}\times\mathcal{K}\rightarrow\mathbb{R}$ is the
reward function for one execution-length decision,
$F_{\mathrm{ex}}:\mathcal{X}\times\mathcal{K}\rightarrow\Delta(\mathbb{N})$ is
the duration distribution over primitive timesteps,
$\rho_{\mathrm{ex}}\in\Delta(\mathcal{X})$ is the initial decision-state
distribution,
and $\gamma$ is the primitive-time discount factor inherited from the underlying
MDP.

The initial decision-state distribution is induced by the environment initial
state distribution and the frozen base policy:
\[
    \begin{gathered}
    s_0\sim\rho_0,\qquad
    \mathbf{a}_0\sim\pi_{\mathrm{base}}(\,\cdot\,\mid s_0),
    \\
    x_0=(s_0,\mathbf{a}_0)\sim\rho_{\mathrm{ex}}.
    \end{gathered}
\]
Given a decision state $x=(s_t,\mathbf{a}_t)$ and an execution length $k_t$, the
environment executes the prefix of the chunk for $k_t$ primitive timesteps:
\[
    a_t,a_{t+1},\ldots,a_{t+k_t-1}.
\]
This produces a trajectory segment
$(s_t,a_t,r_t,\ldots,s_{t+k_t})$ under the primitive dynamics $P$. The reward of
the high-level decision is the discounted reward accumulated along this segment,
\[
    r_{\mathrm{ex}}(x,k_t)
    =
    \mathbb{E}
    \left[
        \sum_{i=0}^{k_t-1}\gamma^i r(s_{t+i},a_{t+i})
        \,\middle|\,
        x,k_t
    \right].
\]
After the prefix execution ends, the base policy is queried again at the new
state $s_{t+k_t}$ to produce the next chunk
$\mathbf{a}'\sim\pi_{\mathrm{base}}(\,\cdot\,\mid s_{t+k_t})$. Therefore, the
next high-level decision state is
\[
    x'=(s_{t+k_t},\mathbf{a}').
\]
Let $K$ denote the realized duration of executing length $k_t$. Starting from
$x=(s_t,\mathbf{a}_t)$, the primitive dynamics generate states
$s_{t+1},\ldots,s_{t+K}$ by executing the prefix of $\mathbf{a}_t$ until either
$k_t$ actions have been executed or the episode terminates. The duration
distribution is
\[
    \begin{aligned}
    &F_{\mathrm{ex}}(K\mid x,k_t)\\
    &\quad=\Pr\!\left(\begin{gathered}
      K\text{ primitive timesteps elapse}\\
      \text{before the next decision}
    \end{gathered}\,\middle|\,x,k_t\right).
    \end{aligned}
\]
In the non-terminal case, $F_{\mathrm{ex}}(K\mid x,k_t)=\mathbbm{1}\{K=k_t\}$;
if the episode terminates early, then $K<k_t$ is the actual number of primitive
steps executed and no bootstrap term is used.

The transition kernel over decision states is
\[
    \begin{aligned}
    &P_{\mathrm{ex}}(x'\mid x,k_t)\\
    &\quad=
    \Pr\!\left(
        s_{t+K}=s',\;
        \mathbf{a}'\sim\pi_{\mathrm{base}}(\,\cdot\,\mid s')
        \mid x,k_t
    \right),
    \\
    &\qquad x'=(s',\mathbf{a}').
    \end{aligned}
\]
That is, $P_{\mathrm{ex}}$ is induced by first rolling out the primitive
dynamics under the selected prefix, and then querying the frozen base policy at
the resulting state to form the next decision state.

For an execution-length policy
$\pi_{\mathrm{ex}}(k\mid x)$, the SMDP Bellman equation is
\[
    Q^{\pi_{\mathrm{ex}}}(x,k)
    =
    \mathbb{E}
    \left[
        R
        +
        \gamma^{K}
        V^{\pi_{\mathrm{ex}}}(x')
        \,\middle|\,
        x,k
    \right],
\]
where
\[
    \begin{gathered}
    R=\sum_{i=0}^{K-1}\gamma^i r(s_{t+i},a_{t+i}),\\
    V^{\pi_{\mathrm{ex}}}(x')
    =
    \mathbb{E}_{k'\sim\pi_{\mathrm{ex}}(\cdot\mid x')}
    \left[
        Q^{\pi_{\mathrm{ex}}}(x',k')
    \right],
    \end{gathered}
\]
and $K$ is the realized duration of the high-level execution-length decision.
When the episode does not terminate during the prefix execution, $K=k$.

This recovers the duration-aware TD target used in ExRL. Writing
$x=(s_t,a_{t:t+H})$, $x'=(s',a')$, and $K=k_t$, the target becomes
\[
    y_t
    =
    \sum_{i=0}^{k_t-1}\gamma^i r(s_{t+i},a_{t+i})
    +
    \gamma^{k_t}
    V(s',a'),
\]
where $V$ is instantiated by the particular RL algorithm, such as a
greedy value for ExRL-Q or a soft value for ExRL-SAC. Thus, the only difference
from a standard one-step Bellman backup is that the selected high-level action
$k_t$ determines the duration of the transition. As a result, $k_t$ affects not
only the accumulated reward and next decision state, but also the bootstrap
discount through the factor $\gamma^{k_t}$.

\section{AI Use Statement}
In this work, we used generative AI tools to assist with code implementation, running experiments, and revising the manuscript. The human authors conceived the research ideas, designed and iteratively refined the experiments, and wrote the initial manuscript, including the development and organization of the related work. We have reviewed all AI-assisted work. In particular, the main author carefully verified every experiment and its results. We take responsibility for the final content of this work, including text, claims or artifacts produced with the aid of generative AI.

\section{Reproducibility Statement}
Algorithms~\ref{alg:exrl_q_full} and~\ref{alg:exrl_sac_full} provide pseudocode for ExRL-Q and ExRL-SAC. Appendix~\ref{sec:experimental_details} documents the tasks, demonstrations, policy architectures, pretraining settings, and adaptation hyperparameters for our experiments. Our adaptations of the official baseline implementations and their hyperparameter selection are detailed in Appendix~\ref{sec:baseline_settings}. Our \href{https://github.com/Frankluox/ExRL}{code release} includes DP and ACT pretraining, ExRL training and evaluation, and the online BC and demonstration-augmented variants on Robomimic, together with installation instructions and task-specific configurations. The \href{https://exrl-project.github.io}{project website} provides simulation and real-world rollouts with execution-length visualizations, as well as an uncut real-robot training recording.

\end{document}